\documentclass[10pt,twocolumn,letterpaper]{article}

\usepackage[pagenumbers]{iccv} 

\definecolor{iccvblue}{rgb}{0.21,0.49,0.74}
\definecolor{customgreen}{rgb}{0.302,0.553,0.349} 
\usepackage[pagebackref,breaklinks,colorlinks,allcolors=iccvblue]{hyperref}
\usepackage{algorithm}
\usepackage{algpseudocode}
\usepackage{amsmath}
\usepackage{makecell}
\usepackage[flushleft]{threeparttable}
\usepackage{caption}
\usepackage{multirow}
\usepackage{arydshln}
\usepackage{amsmath}
\usepackage{colortbl}
\usepackage{xcolor}  
\usepackage{pifont}
\usepackage{tabularx} 

\def\paperID{5634} 
\def\confName{ICCV}
\def\confYear{2025}

\title{Sat2City: 3D City Generation from A Single Satellite Image with Cascaded Latent Diffusion}

\author{Tongyan Hua$^{1}$ \quad 
Lutao Jiang $^{1}$ \quad 
Ying-Cong Chen $^{1}$$^{,2}$ \quad 
Wufan Zhao$^{1}$\thanks{Corresponding author.}\\
$^{1}$HKUST(GZ)  \quad $^{2}$HKUST\\
{\tt\small  \{thua388, ljiang553\}@connect.hkust-gz.edu.cn,
yingcongchen@ust.hk,
wufanzhao@hkust-gz.edu.cn}
\\
\small{Project Page: \url{https://ai4city-hkust.github.io/Sat2City/}}
}

\begin{document}
\maketitle

\begin{abstract}
Recent advancements in generative models have enabled 3D urban scene generation from satellite imagery, unlocking promising applications in gaming, digital twins, and beyond.
However, most existing methods rely heavily on neural rendering techniques, which hinder their ability to produce detailed 3D structures on a broader scale, largely due to the inherent structural ambiguity derived from relatively limited 2D observations.
To address this challenge, we propose \textbf{Sat2City}, a novel framework that synergizes the representational capacity of sparse voxel grids with latent diffusion models, tailored specifically for our novel 3D city dataset. Our approach is enabled by three key components: 
(1) A cascaded latent diffusion framework that progressively recovers 3D city structures from satellite imagery, (2) a Re-Hash operation at its Variational Autoencoder (VAE) bottleneck to compute multi-scale feature grids for stable appearance optimization and (3) an inverse sampling strategy enabling implicit supervision for smooth appearance transitioning.
To overcome the challenge of collecting real-world city-scale 3D models with high-quality geometry and appearance, we introduce a dataset of synthesized large-scale 3D cities paired with satellite-view height maps. Validated on this dataset, our framework generates detailed 3D structures from a single satellite image, achieving superior fidelity compared to existing city generation models.
\end{abstract}

\begin{figure}[t!]
  \centering
   \includegraphics[width=1\linewidth]{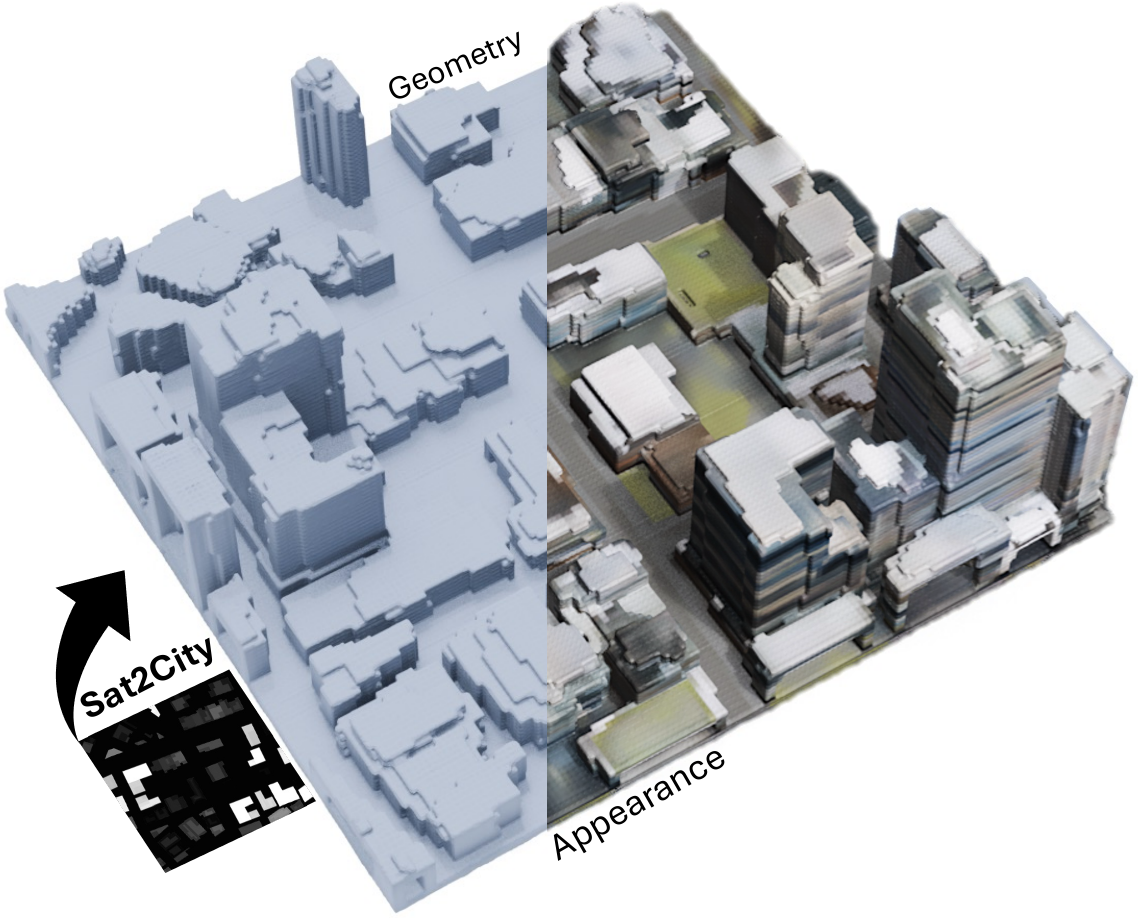}

   \caption{We present \textbf{Sat2City}, a novel framework for generating high-fidelity 3D city models with detailed geometry and appearance from a single satellite observation, in \textbf{\(\sim\) 1 minute}, using cascaded latent diffusion. This process is achieved even without the need for auxiliary inputs like segmentation map.}
   \label{fig:onecol}
\end{figure}

\section{Introduction}
\label{sec:intro}


Generating realistic 3D urban scenes~\cite{city_gen-li2024sat2scene,city_gen-lin2023infinicity,city_gen-xie2024citydreamer,city_gen-xie2024gaussiancity} has gained significant attention in recent years, driven by applications in gaming, urban planning, and digital twin systems. With advancements in neural rendering techniques, most prior works have focused on generating novel street-view images or videos~\cite{vid_img-li2021sat2vid,vid_img-lu2020geometry,vid_img-qian2023sat2density,vid_img-shi2022geometry,vid_img-xu2024geospecificviewgeneration,vid_img-li2024crossviewdiff,vid_img-deng2024streetscapes,vid_img-lu2024urban,vid_img-yang2023urbangiraffe,vid_img-li2024syntheocc} within urban environments. 
These methods are restricted to rendering cities from a highly limited set of viewpoints and trajectories, and the generated images or videos often lead to poor results when explicitly reconstructing a 3D scene.

%



Recent methods have evolved to integrate more reliable priors, such as segmentation and height maps from satellite imagery, enabling a more salient abstraction of urban environments and extending novel view synthesis to broader scales.
Early on, InfiniCity~\cite{city_gen-lin2023infinicity} directly uses lifted voxels from satellite imagery as feature volumes of neural radiance fields (NeRFs), training a generative adversarial network (GAN) to transform ray-sampled features into rendered pixels. Building upon this framework, CityDreamer~\cite{city_gen-xie2024citydreamer} and its Gaussian variants~\cite{city_gen-xie2024gaussiancity} achieve more realistic rendering effects by implementing feature-voxel partitioning through per-pixel semantic annotation. However, these 2D rendering-based generation approaches still suffer from significant 3D ambiguity due to the absence of direct supervision from 3D textured coordinates, leading to limited capability for reconstructing detailed 3D structures.
Recently, Sat2Scene~\cite{city_gen-li2024sat2scene} pioneers the use of diffusion models for direct color generation on 3D point clouds. While this constitutes a conceptual advance, the method faces practical limitations: The color generation process relies on a densely allocated point cloud constructed around predefined geometry derived from satellite height maps. Despite achieving photorealistic appearance rendering, this process lacks the capability to refine the geometry and, more importantly, is computationally expensive to scale. Furthermore, its applicability is constrained by the scarcity of high-quality, textured point cloud data at urban scales, limiting its usage to single street-level scenarios.

To address these challenges, we introduce \textbf{Sat2City}, a novel generative framework that jointly synthesizes geometry and appearance. Our model learns directly from our high-quality 3D city dataset represented as colorized point clouds, ensuring a unified representation for structured 3D city generation.
Recent advances in 3D generative modeling have demonstrated significant progress in scalability with XCube~\cite{ren2024xcube}, an efficient framework that combines sparse voxel grids and latent diffusion models for large-scale outdoor scene synthesis. At its core, XCube leverages a compact VDB data structure, enabling fast querying and logarithmic memory scaling while preserving expressive power.
Building on this foundation, we adopt XCube’s sparse voxel grid as a unified neural representation for joint geometry and appearance encoding. In contrast to its closed-form successor, SCube~\cite{ren2024scube}, which employs Gaussian Splatting~\cite{kerbl20233d} with image-space texture priors, our framework directly encodes appeance as voxel color attributes. 
Rather than introducing excessive complexity, we leverage native sparse voxels to streamline the pipeline while ensuring consistent 3D geometry and appearance, free from multi-view reprojection artifacts. 

%
%
We identify two crucial findings for capacitating the sparse latent grids to coherently encode the appearance. 
First, performing multi-level coarsening, dubbed Re-Hash, at the VAE bottleneck plays an important role not only in facilitating stable optimization, but also in providing global context for smooth appearance encoding. Second, supervising vertex color attribute learning implicitly through inverse sampling at input point cloud level could enhance smooth visual transitions. 
In order to recover the geometric distribution of the original 3D data—onto which our appearance is assigned—from height maps that heavily perturbed by noise, we additionally train a VAE with a densified bottleneck to encode geometry. By conditioning the sampling of this dense latent volume on elevated height maps, our cascaded latent diffusion model can progressively correct, refine, and colorize the geometry derived from noisy height maps through triple-level latent space conditioning, where each level is conditioned on the previous one.
%
%
%

In summary, our primary contributions are as follows:
\begin{itemize}
\item We present the first 3D city generation framework to achieve city-scale appearance modeling with explicit geometry, producing high-fidelity and controllable urban environments from a single satellite image (\Cref{tab:comp-baseline}).
\item We propose three key innovations enabling joint appearance and geometry generation with sparse voxel grids and latent diffusion: (1) Re-Hash, (2) Inverse Sampling, and (3) a triplet cascaded latent diffusion framework.
\item We introduce a new dataset of orthorectified satellite images paired with 3D colorized point clouds, facilitating training of our framework.
\end{itemize}

\begin{table}[t!]
\centering
\setlength{\tabcolsep}{0.5mm}{
\scalebox{0.9}{
\begin{tabular}{cccc}
\toprule
 & City-Scale & Appearance & Explicit 3D \\
\midrule
XCube~\cite{ren2024xcube} & \textcolor{red}{\ding{55}}& \textcolor{red}{\ding{55}}& \textcolor{green}{\ding{51}}\\
SCube~\cite{ren2024scube} & \textcolor{red}{\ding{55}}& \textcolor{green}{\ding{51}}& \textcolor{green}{\ding{51}}\\
CityDreamer~\cite{city_gen-xie2024citydreamer} & \textcolor{green}{\ding{51}} & \textcolor{green}{\ding{51}} & \textcolor{red}{\ding{55}}\\
Sat2Scene~\cite{city_gen-li2024sat2scene} &\textcolor{red}{\ding{55}}  &\textcolor{green}{\ding{51}}  & \textcolor{green}{\ding{51}}\\
Ours & \textcolor{green}{\ding{51}}& \textcolor{green}{\ding{51}}& \textcolor{green}{\ding{51}}\\

\bottomrule
\end{tabular}}
}
\vspace{-8pt}
\captionsetup{justification=justified}
\caption{Comparison of 3D urban scene generation methods.}
\vspace{-12pt}
\label{tab:comp-baseline}
\end{table}
\section{Related Works}
\label{sec:relate}
\textbf{3D Generation in Object-level.}
The recent 3D generation methods can be divided into two paradigms: per-scene optimization and direct generation.
Per-scene optimization methods~\cite{poole2022dreamfusion, bai2023componerf, zhou2024gala3d, cohen2023set, yuan2024dreamscape, jiang2024general, epstein2024disentangled, cheng2023progressive3d} typically require several hours of training and are limited to specific scenes, making them less flexible.
In contrast, we opt to follow the direct generation paradigm, which offers greater efficiency.
The direct generation can be further categorized into three types, including feed-forward~\cite{hong2023lrm, xu2024instantmesh}, diffusion~\cite{zhang2024clay, ren2024xcube, ren2024scube, zhang20233dshape2vecset}, and auto-regressive~\cite{siddiqui2024meshgpt, chen2024meshanything, chen2024meshanything2, chen2024meshxl, tang2024edgerunner, weng2024pivotmesh, wang2024llama}.
Given that feed-forward methods are constrained by limited diversity and auto-regressive methods are not well-suited for large-scale scene generation, we focus on diffusion-based methods due to its merits of the large-scale, diversity, and speed.
In diffusion-based methods, two main representations are commonly used: latent sets~\cite{zhang20233dshape2vecset, zhang2024clay, zhang2024lagem} and sparse voxels~\cite{ren2024xcube, ren2024scube}. We choose the latter one as our scene representation for its inherent computational efficiency and proven superiority in handling large-scale scenes.


\noindent\textbf{3D Generation in Scene-level.}
Two prominent strategies have emerged to specialize scene generation~\cite{sce-fridman2024scenescape,sce-xu2024sketch2scene,sce-yang2024scene123,sce-zhang20243d} for city generation: asset retrieval and volumetric neural rendering. The first strategy focuses on agent-based procedural retrieval of pre-built 3D assets, such as those available in Blender, to enable the automated planning of 3D cities~\cite{city_agent-deng2024citycraft,city_agent-shang2024urbanworld,city_agent-yang2024procedural,city_agent-zhang2024cityx} or more generalized scenes~\cite{city/scene_agent-zhou2024scenex,liu2025worldcraft}. 
However, this approach is inherently limited by the existing asset library, which makes it challenging to generate novel content. In contrast, neural rendering techniques focus on generating novel street-view images or videos
~\cite{sce-chai2023persistent,sce-chen2023scenedreamer,sce-fridman2024scenescape,sce-hao2021gancraft,sce-xu2024sketch2scene,sce-yang2024scene123,sce-zhang20243d,vid_img-deng2024streetscapes,vid_img-yang2023urbangiraffe,vid_img-lu2024urban,vid_img-li2024syntheocc}. 
Recent advancements have significantly improved cross-view consistency over long trajectories, particularly when leveraging large datasets~\cite{vid_img-deng2024streetscapes}. 
More recent work has explored the use of salient urban scene representations, such as semantic volumes~\cite{vid_img-lu2024urban,vid_img-yang2023urbangiraffe,vid_img-li2024syntheocc}, as controllable priors for stable city-scale navigation. Yet, the inherent ambiguity of 3D structures inferred from 2D generative priors continues to constrain the robustness of these approaches when handling diverse viewpoints. 
Multi-view consistency has been significantly improved by separately modeling elements in urban environments with fine-grained categories~\cite{city_gen-lin2023infinicity,city_gen-xie2024citydreamer,city_gen-xie2024gaussiancity}, though 3D consistency remains suboptimal. Training directly on 3D data presents a promising opportunity; however, existing methods are still constrained to street-level generation and have not scaled to entire cities~\cite{city_gen-li2024sat2scene,ren2024scube,lu2024infinicube}. In contrast, our approach enables city-level generation by directly learning from large-scale 3D urban data, ensuring both geometric and appearance consistency.

\section{Method}
\label{sec:method}

\begin{figure}[t!]
  \centering
   \includegraphics[width=1\linewidth]{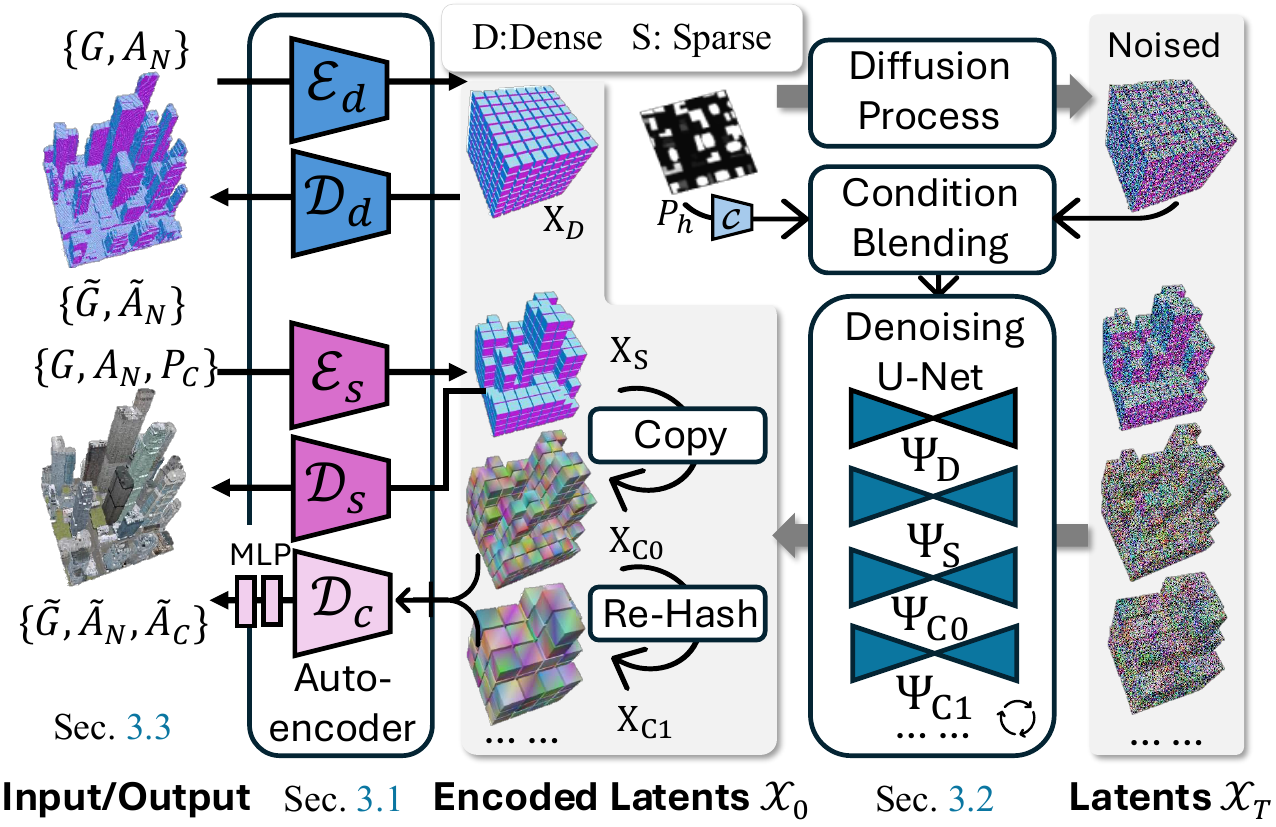}

   \caption{Sat2City Training Pipeline.}
   \label{fig:framework}
   \vspace{-12pt}

\end{figure}

Our method is trained on 3D city data, represented as a colorized point cloud $P_C \in \mathbb{R}^{N \times 6}$, along with corresponding height maps, which are also elevated to a point cloud $P_h$ (see \Cref{fig:dataset-hmap}). 
For geometry encoding, the point clouds are voxelized into sparse voxel grids $G$, with each vertex assigned trilinearly splatted normal attributes $A_{N}$ from the point cloud. For appearance encoding, the per-vertex color attribute $\tilde{A}_C$ is implicitly derived from the colorized point cloud during training via trilinear interpolation at $P_C$.

As depicted in \Cref{fig:framework}, our architecture integrates multiple VAEs with encoder-decoder pairs $(\mathcal{E}, \mathcal{D})$ and latent diffusion U-Nets $\Psi$ through a two-stage training paradigm. The VAEs initially encode sparse voxel grids into compact latent representations $\mathcal{X}_0 = \{X_D, X_S, \{X_{Ck}\}_{k=0}^n\}$, where $X_D$ and $X_S$ respectively capture dense and sparse geometric features while $\{X_{Ck}\}_{k=0}^n$ encodes hierarchical appearance details through $n$ progressively coarsen levels (\Cref{method:vae}). Following this encoding phase, the latent diffusion model undergoes optimization to iteratively denoise Gaussian-distributed features $\mathcal{X}_T$ across $T$ timesteps, progressively recovering the target distribution $\mathcal{X}_0$ (\Cref{method:diff}). The framework is ultimately validated on our proposed dataset detailed in \Cref{method:data}.

%

\subsection{Triplet Bottleneck VAE}
\label{method:vae}

\begin{figure}[t!]
  \centering
   \includegraphics[width=1\linewidth]{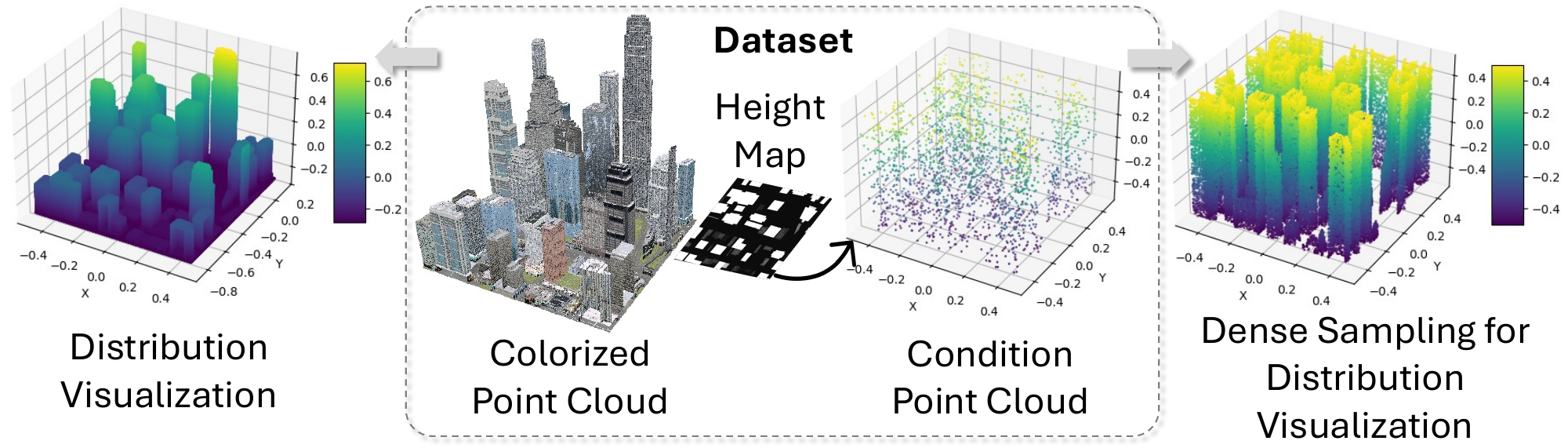}
\caption{\textbf{Raw Input Data.} The colorized point cloud $P_C$ and height map-derived conditional point cloud $P_h$ are normalized for visualization, with $P_h$ upsampled to enhance distribution clarity.}
   \label{fig:dataset-hmap}
   \vspace{-12pt}
\end{figure}

\begin{figure}[t!]
  \centering
   \includegraphics[width=0.92\linewidth]{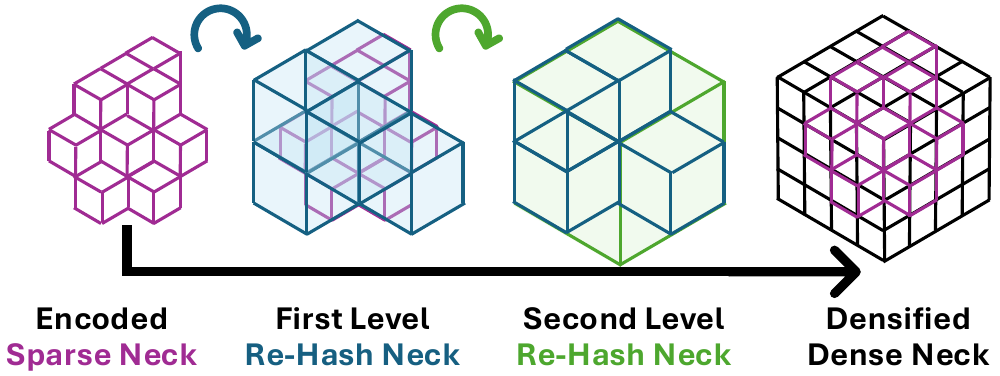}
\caption{\textbf{Triplet Bottleneck Architecture for VAE.} All VAEs first downsample the input sparse voxel grid into an \textbf{Encoded Sparse Neck}, directly decoded by $\mathcal{D}_s$, while undergoing densification into a \textbf{Dense Neck} for $\mathcal{D}_d$ and re-hashing into a multi-level \textbf{Re-Hash Neck} for $\mathcal{D}_c$.}
   \label{fig:tri-neck}
   \vspace{-12pt}
\end{figure}

Our VAEs utilize sparse convolutional neural networks as encoders to downsample the input sparse voxel grid. It then employs a structure prediction backbone~\cite{huang2023neural} as a decoder, which iteratively subdivides existing voxels and prunes excessive ones, ultimately upsampling back to the same resolution as the input~\cite{ren2024xcube}. 
As illustrated in \Cref{fig:tri-neck}, instead of maintaining identical sparse latent grids for all VAE bottlenecks, we introduce two additional structures, namely the dense neck and the re-hash neck, to represent the overall occupancy and appearance field for downstream diffusion sampling, respectively.
The primary motivation behind densification is to enable diffusion models to explicitly distinguish between occupied and unoccupied volumes, thereby eliminating falsely allocated volumes introduced by conditional inputs, which are highly perturbed by noise (\Cref{fig:dataset-hmap}). Additionally, implementing densification in the bottleneck rather than in the input structure sustains an affordable computational cost for large-scale scene encoding.

\noindent\textbf{Re-Hash Neck.}
Although dense volumetric representations are recommended for preserving detailed structures when modeling 3D shapes in various tasks~\cite{hua2024benchmarking,hui2022neural}, multi-level encoding, which is essential for smooth appearance transitions~\cite{johari2023eslam,wang2023co}, imposes a substantial computational burden when applied to scene-level data encoding~\cite{zhu2022nice}.
Therefore, we introduce Re-Hash, a hierarchical coarsening mechanism that progressively restructures the voxel representation through iterative resampling. Given an initial voxel grid \(X_{C0}\), derived from sparse geometric feature grids via direct ``copy'' (\Cref{fig:framework}), we construct a multi-depth hierarchy in which each subsequent level is coarsened by a factor of two in voxel size (\Cref{fig:tri-neck}). Formally, for each depth \(n\), the voxel resolution is updated as: $v_n = 2^n v_0, \quad o_n = \frac{v_n}{2}$, where \( v_n \) and \( o_n \) denote the voxel size and origin at depth \( n \), respectively. 
%
%
At each level, the new sparse feature grid \(X_{Cn} \) is assigned with the features from \(X_{C{n-1}} \) using trilinear interpolation, ensuring smooth transitions between resolutions. 
%
%
Specifically, given a set of 3D query positions from the restructured grid $G_{Cn}$, we sample latent features from the previous depth using trilinear interpolation (Tri):
{\setlength\abovedisplayskip{2pt}
\setlength\belowdisplayskip{2pt}
\begin{equation}
X_{Cn} = \text{Tri}(G_{Cn},X_{Cn-1}),
\end{equation}}where $X_{Cn}$ represents the encoded features at depth \( n \). The final Re-Hash hierarchy preserves critical appearance details at multiple scales while facilitating a progressively refined latent space for downstream processing.


%
%
%
%

\noindent\textbf{Dual-Stage Appearance Training.} We train two independent VAEs for dense and sparse scenarios, respectively. 
For dense geometry, we employ a conventional VAE, consisting of an encoder and a decoder. 
Whereas in the sparse scenario, our architecture employs a unified encoder $\mathcal{E}_s$ to approximate the posterior distributions of the sparse geometric latent variable $X_S$ and the multi-level appearance latent variable $X_{Cn}$. The model then utilizes two distinct decoders: the sparse decoder $\mathcal{D}_s$ reconstructs geometry and normal attributes via the conditional likelihood $p_{\mathcal{D}_s}(G, A_N | X_S)$, while the appearance decoder $\mathcal{D}_c$ processes each level of the re-hashed feature grids, concatenates them, and feeds the result into a small multilayer perceptron (MLP) to synthesize appearance, as demonstrated in~\Cref{fig:framework} and detailed in the following paragraph.
%
%
Notably, the sparse latent $X_S$ serves dual purposes: it provides structural guidance for progressive grid pruning in $X_{Cn}$ during training, while also eliminating unnecessary voxel grids that were not suppressed during dense neck pruning.
%
%
The training of the sparse VAE begins by exclusively optimizing $q_{\mathcal{E}_s}(X_S | G, A_N)$ and $p_{\mathcal{D}_s}(G, A_N | X_S)$ for geometric fidelity until epoch $E$. After this stage, $X_S$ initializes the finest-level appearance latent volume $X_{C0}$, which is subsequently integrated with re-hashed coarser grids and processed through $\mathcal{D}_c$ to constrain the appearance latent $X_{Cn}$.

\begin{figure}[t!]
  \centering
   \includegraphics[width=1\linewidth]{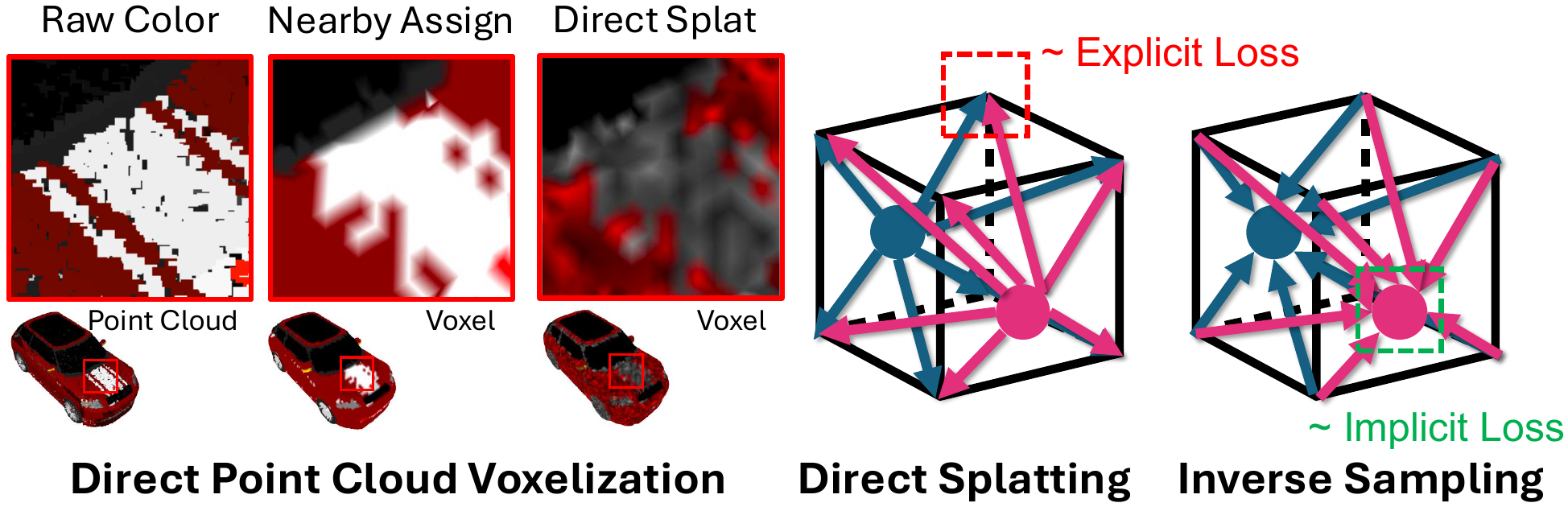}
   \caption{\textbf{Direct Point Cloud Voxelization and Inverse Sampling Process.} Examples of challenging scenarios in per-vertex color attribute learning, illustrated using a dense point cloud generated from the ShapeNet dataset~\cite{chang2015shapenet} for the easy of demonstration.}
   \label{fig:inverse-sample}
   \vspace{-12pt}
\end{figure}

\noindent\textbf{Inverse Sampling.}
Learning on 3D datasets provides generative models with significant advantages in terms of speed and spatial consistency. Moreover, for appearance learning, it offers direct supervision, reducing ambiguity. However, directly performing per-vertex color attribute learning is challenging. 
As illustrated in \Cref{fig:inverse-sample}, voxelizing colorized point clouds presents challenges with both naive strategies. (1) Direct assignment from the nearest point results in a lack of smoothness. (2) Trilinear splatting causes color blending conflicts due to overlapping contributions from multiple points.
To address these issues, this work constrains \(A_N\) only at the vertex level while implicitly guiding the learning of per-vertex color attributes \(A_C\) through inverse sampling, a strategy inspired by the sampling process in NeRF's voxel grid variants~\cite{sun2022direct,mildenhall2021nerf,zhu2022nice}.
During the training stage, for each level of the appearance latent hierarchy \(X_{Cn}\), it is separately decoded through \(\mathcal{D}_c\) into per-vertex color features, which are then trilinearly sampled on the colorized point cloud \({P}_C\). The interpolated features from all levels are concatenated and processed by an MLP, resulting in estimated point cloud color $\tilde{P}_C$:
{\setlength\abovedisplayskip{2pt}
\setlength\belowdisplayskip{2pt}
\begin{equation}
\tilde{P}_C = \text{MLP}\{\oplus^{n}_{k=0} 
Tri\left(
P_C,\mathcal{D}_c(X_{Cn})
\right)
\},
\label{eq:mlp}
\end{equation}}where \(\oplus\) denotes the concatenation operation applied to the decoder outputs. 
At inference time, given the predicted city geometry in the form of grid vertices \(\tilde{G}\), we can infer its color in the same manner:
{\setlength\abovedisplayskip{2pt}
\setlength\belowdisplayskip{2pt}
\begin{equation}
\tilde{A}_C = \text{MLP}\{\oplus^{n}_{k=0} 
Tri\left(
\tilde{G},\mathcal{D}_c ( X_{Cn} )
\right)
\}.
\label{eq:mlp}
\end{equation}}




\subsection{Conditional Cascaded 3D Latent Diffusion}
\label{method:diff}
\begin{figure}[t!]
  \centering
   \includegraphics[width=1\linewidth]{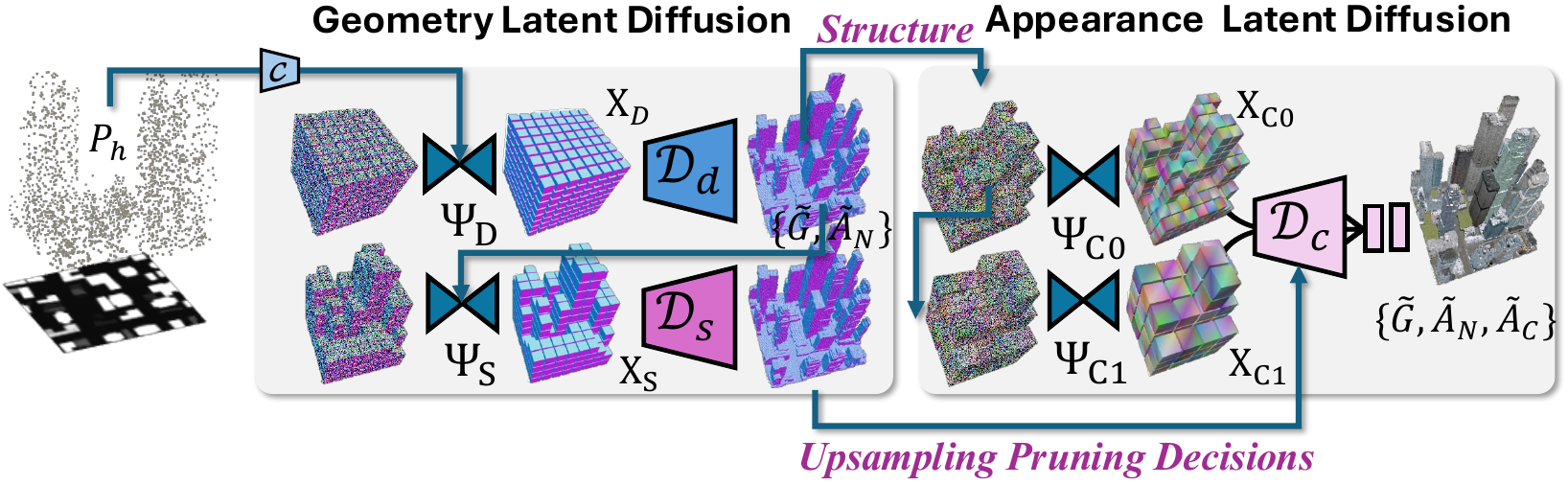}
\caption{Inference pipeline of Sat2City.}
   \label{fig:infer}
   \vspace{-12pt}
\end{figure}

Diffusion models are probabilistic frameworks designed to model target data distributions, such as the latent space \( \mathcal{X}_0 \sim q(\mathcal{X}_0) \) in VAEs (\ie latent diffusion model), by progressively denoising a randomly sampled variable \( \mathcal{X}_T \sim \mathcal{N}(0, I) \), initially drawn from a Gaussian distribution~\cite{rombach2022high}. The process is based on learning the reverse denoising procedure of a fixed Markov Chain with \( T \) steps. Training starts with the diffusion process, which progressively adds noise to \( \mathcal{X}_0 \) over \( T \) steps to form \( \mathcal{X}_T \). The model then iteratively denoises \( \mathcal{X}_T \) using U-Nets, gradually recovering the original latent representation \( \mathcal{X}_0 \) (\Cref{fig:framework}).
However, single-stage latent diffusion is proven to be inadequate when addressing expansive 3D landscapes~\cite{ren2024xcube}, especially when the appearance is also required~\cite{ren2024scube,lu2024infinicube}.
%
%
%
Our model therefore adopts a sequentially conditioned diffusion pipeline, where later stages incorporate contents from earlier stages, as illustrated in~\Cref{fig:infer}. Our generation process comprises three consecutive sampling stages for dense geometry latent \(X_D\), sparse geometry latent \(X_S\), and multi-level sparse appearance latent \(X_{Cn}\), using DDIM~\cite{song2020ddim}. The process begins by formulating the joint distribution factorization with respect to the dense grid and its corresponding latent representations as follows:
{\setlength\abovedisplayskip{2pt}
\setlength\belowdisplayskip{2pt}
\begin{equation}
\begin{array}{ll}
p(X_D, G, A_N) = p_{\mathcal{D}_d}(G, A_N | X_D) p_{\Psi_D}(X_D | c(P_h)).
\end{array}
\end{equation}}Here, the initial dense geometry latent \(X_D\) encodes the overall spatial layout based on the encoded conditional height field inputs \(P_h\). During inference, the denoised dense latent is decoded by \(\mathcal{D}_d\) to generate the sparse grid output \(\{G, A_N\}\), which facilitates sparse latent diffusion. This intermediate representation serves as a bridge between geometry and appearance, leveraging \(\{G, A_N\}\) to fit a sparse latent volume for finer surface reconstruction. Crucially, the sparse latent grid decoder records voxel pruning decisions, \textit{struct}, at each upsampling step, guiding structured pruning during appearance decoding, since its re-hashed latent grid does not explicitly encode geometric distributions.

{\setlength\abovedisplayskip{2pt}
\setlength\belowdisplayskip{2pt}
\begin{equation}
\begin{array}{ll}
p(X_S,  \textit{struct}) = p_{\mathcal{D}_s}(G, A_N | X_S) p_{\Psi_S}(X_S | G,A_N).
\end{array}
\end{equation}}\textit{struct} are subsequently leveraged to guide the appearance upsampling process, ensuring that the finest-level appearance feature grid maintains a consistent geometric structure with sparse latents at each upsampling step of the appearance decoder:
{\setlength\abovedisplayskip{2pt}
\setlength\belowdisplayskip{2pt}
\begin{equation}
\begin{array}{ll}
p(G, A_N, A_C) = p_{\mathcal{D}_c}(G, A_N, A_C | X_{C}) \cdot \\
\qquad \qquad \qquad \quad \prod_{n=0}^{N} p_{\Psi_{Cn}}(X_{Cn} | \textit{struct}).  
\end{array}
\end{equation}}Specifically, the finest-level appearance feature grid $X_{C0}$ undergoes a \textit{re-hashing} process, generating multiple coarser levels. During each upsampling step, these coarser levels (\eg, $X_{C1}$ in~\Cref{fig:infer}) are iteratively re-fitted by adapting to the pruned voxel structure defined by $X_{C0}$. 
Refer to \textit{Supplementary Material} for more details on technical implementations.
%
%
%

\begin{figure}[t!]
  \centering
   \includegraphics[width=0.92\linewidth]{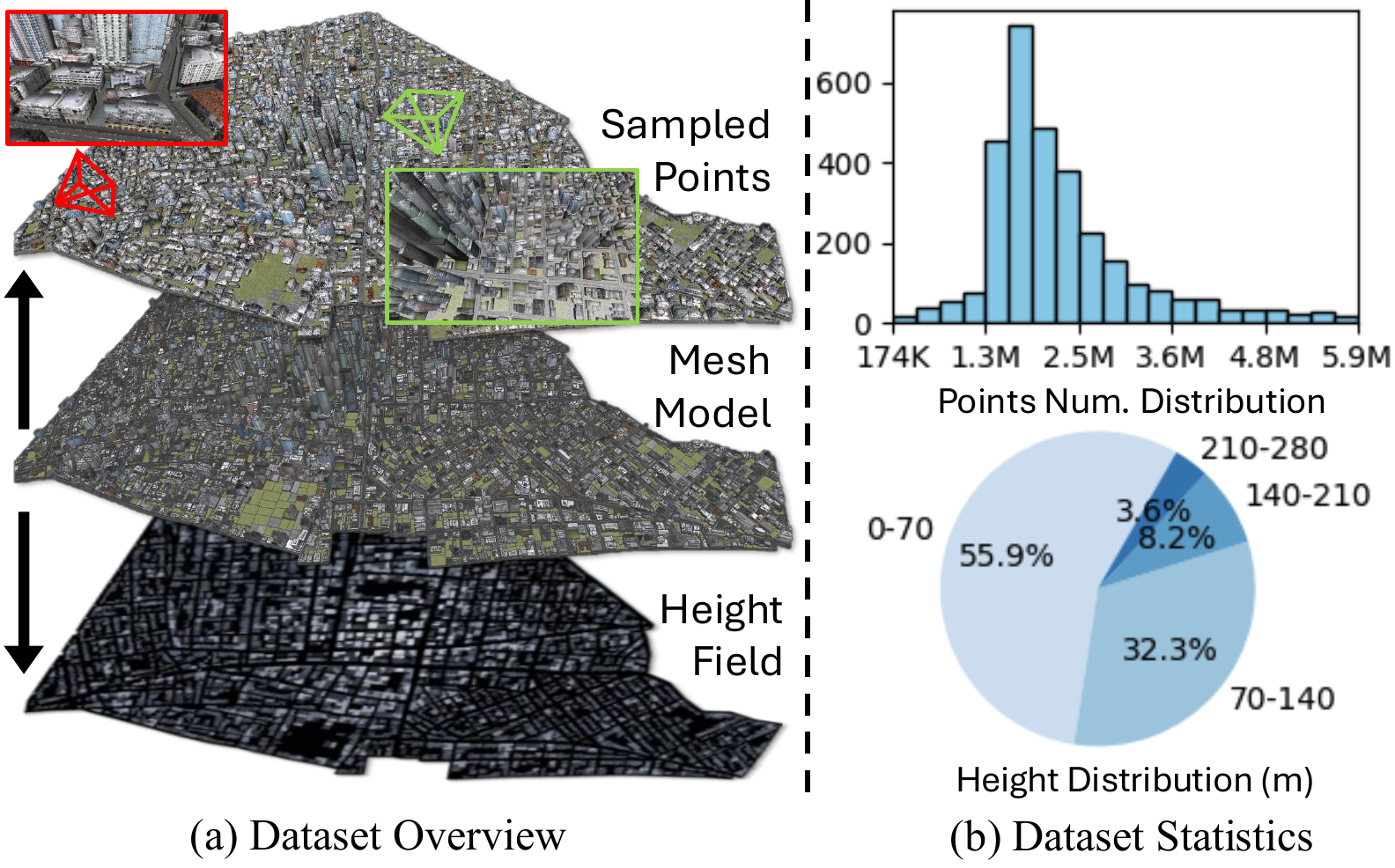}
   \caption{Overview of the Sat2City dataset.}
   \label{fig:dataset}
   \vspace{-12pt}
\end{figure}

\subsection{Sat2City Dataset}
\label{method:data}
The 3D city dataset is created by artists in Blender~\cite{blender} as mesh-based models. To generate the point cloud, we randomly sample 100 million points from the mesh surfaces using CloudCompare~\cite{cloudcompare}.
The height field is then simulated in blender by mapping the texture coordinates of the elevation axis (in this case, the y-axis) into grayscale, with the lowest and highest elevations being linearly assigned to 0 and 255, respectively. 
We then position a vertically aligned orthographic camera to render a high-resolution synthetic height map of the entire city in a single shot. The output image, at a resolution of $2268 \times 3423$ pixels, covers a land area of $2090 \times 3449.4$ \(m^2\), achieving a ground sampling distance of 0.93 \(m^2\) per pixel.
To enhance realism and align our simulation with real-world height maps—commonly referred to as digital surface models (DSMs) in remote sensing and photogrammetry—the height field rendering incorporates ambient lighting variations. Additionally, noise is introduced by applying contrast scaling to the rendered height map (\Cref{fig:dataset-hmap}).

\Cref{fig:dataset} presents a layered visualization of the raw dataset alongside its statistical insights. In~\Cref{fig:dataset} (a), the mesh model represents the original 3D city design in a mesh form, with the corresponding height field displayed in the bottom layer and the sampled point cloud in the top layer.
The raw sampled point cloud is first spatially aligned with the height field images, ensuring a one-to-one correspondence. The aligned data is then uniformly cropped into \(300\times300\) pixel segments, preserving the structural consistency between the point cloud and height field throughout the dataset.
The entire dataset is randomly partitioned into 3110 instances, with \(10\%\) reserved for testing and validation, ensuring no data leakage during training. 
An example of a cropped height map and its corresponding point cloud is shown in~\Cref{fig:dataset-hmap}. These cropped point cloud instances contain between 174,000 and 5.9 million points, as visualized in~\Cref{fig:dataset} (b). Please refer to \textit{Supplementary Material} for more details.

%











\begin{figure*}[t!]
  \centering
   \includegraphics[width=1\linewidth]{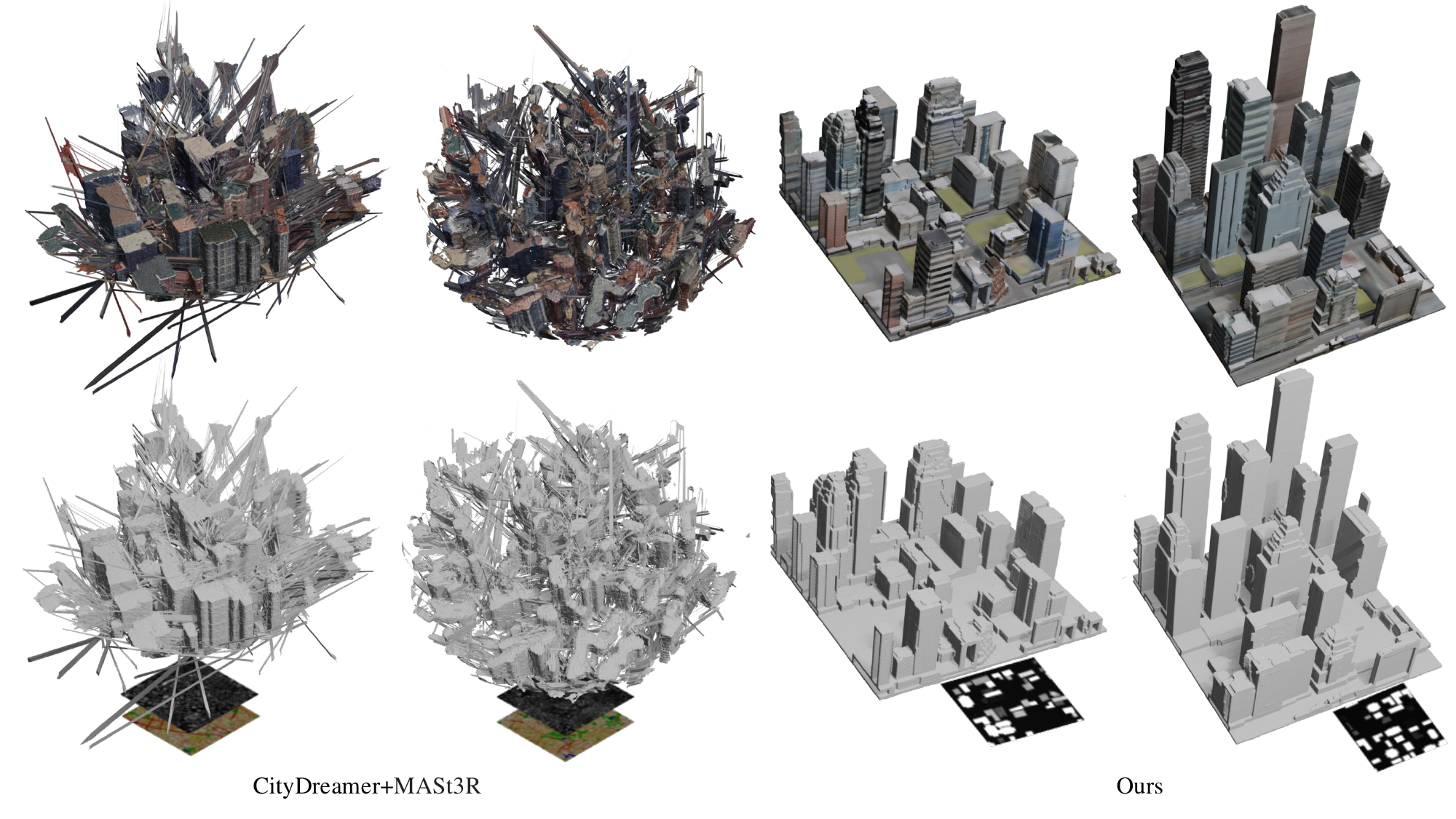}
\caption{\textbf{Qualitative Comparison with CityDreamer~\cite{city_gen-xie2024citydreamer}}. CityDreamer requires both segmentation and height maps, while Sat2City relies solely on height maps. CityDreamer generates 2D city renderings, reconstructed into meshes via MASt3R~\cite{leroy2024grounding}.}
   \label{fig:comapre-citydreamer}
   \vspace{-12pt}
\end{figure*}

\begin{figure}[t!]
  \centering
   \includegraphics[width=0.9\linewidth]{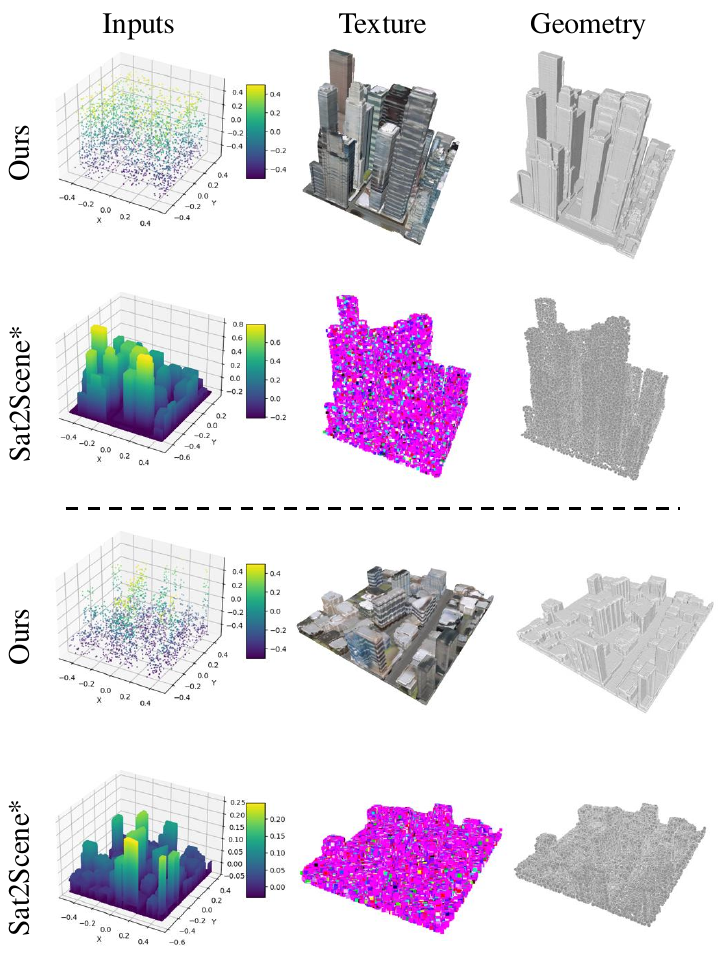}
\caption{\textbf{Qualitative Comparison with Sat2Scene~\cite{city_gen-li2024sat2scene} Trained on Our Dataset}. Since Sat2Scene lacks geometry generation, we provide ground truth point clouds as input, whereas Sat2City generates geometry from noised height maps.}
   \label{fig:comapre-sat2scene}
   \vspace{-12pt}
\end{figure}

\section{Experiments}

\subsection{Evaluation Protocols}

\textbf{Geometric Metrics.} To assess the quality of the generated 3D city models, we extract their mesh representations and compute three widely used metrics for 3D generation~\cite{hui2022neural,yang2019pointflow,siddiqui2024meshgpt,ren2024xcube,wu2024blockfusion}: Coverage score (COV) and Minimum Matching Distance (MMD). 
COV quantifies the proportion of points in the reference point cloud that have at least one corresponding match in the generated point cloud, While MMD computes the average distance by identifying the nearest neighbor in the generated set for each point in the reference point cloud. 
These metrics are evaluated using Chamfer Distance (CD) and Earth Mover’s Distance (EMD), which serve as metrics to quantify the geometric similarity between the generated and reference point clouds. 10,000 points are randomly sampled from both the generated mesh surface and the reference point cloud for metric computation.
Please refer to~\cite{yang2019pointflow} for more details.

\noindent\textbf{Appearance Metric.}
A user study was conducted with 60 participants who evaluated the scene generation results based on two criteria: Perceptual Quality (PQ) and Structural Completeness (SC), using a 10-point scale. Evaluations were performed in both appearance-based mode (TPQ and TSC) and geometry-only mode (GPQ and GSC). In the appearance-based mode, participants assessed a textured mesh, while in the geometry-only mode, the appearance was replaced with a monochromatic material to emphasize the geometric structure~\cite{wu2024blockfusion}. Please refer to the \textit{Supplementary Material} for the survey design.  

%



\begin{table}[t!]
\centering
\setlength{\tabcolsep}{0.4mm}{
\scalebox{0.85}{
\begin{tabular}{lccc|c}
\toprule
 &  \multicolumn{2}{c}{MMD$\downarrow$} & \multicolumn{2}{c}{COV(\%,$\uparrow$)}  \\
\cmidrule(lr){2-3}
\cmidrule(lr){4-5}
& CD & EMD & CD & EMD \\ 
\hline
 
NFD (unconditional)~\cite{shue20233d} & 0.0445& 0.2363& 22.66& 29.66 \\ 
BlockFusion (unconditional)~\cite{wu2024blockfusion} & 0.0326& 0.1865& 50.49& 55.66 \\ 
 \textbf{Ours} (conditional)& \textbf{0.0165}& \textbf{0.1157}& \textbf{100.00}& \textbf{60.00} \\ 
\bottomrule
\end{tabular}}
\vspace{-8pt}
\captionsetup{justification=justified}
\caption{\textbf{Geometric Quality Comparison}. Results for NFD and BlockFusion are obtained through unconditional generation methods~\cite{wu2024blockfusion}, while our Sat2City framework operates as a conditional generation pipeline.}
\vspace{-12pt}
\label{tab:geo_eval}
}
\end{table}

\subsection{Comparison}

\noindent\textbf{Baselines.} To the best of our knowledge, this work presents the first 3D city generation framework trained directly with 3D supervision for both geometry and appearance encoding. In the absence of directly comparable methods, we adopt a two-fold evaluation strategy. For geometry generation quality assessment, we compare against established 3D scene generation pipelines, including BlockFusion~\cite{wu2024blockfusion} and its reproduction of NFD~\cite{shue20233d}.  
For visual quality evaluation, we benchmark against rendering-based urban generation baselines, CityDreamer~\cite{city_gen-xie2024citydreamer} and Sat2Scene~\cite{city_gen-li2024sat2scene}. Notably, Sat2Scene serves as the most relevant baseline, as it is the only architecture compatible with our dataset configuration. Retraining all baselines is challenging, as BlockFusion (watertight meshes) and CityDreamer (aerial images) require inputs incompatible with Sat2City (point clouds). Therefore, to ensure a fair comparison, we faithfully re-implement Sat2Scene on our dataset using its official code.

\noindent\textbf{Qualitative Comparison.}
\Cref{fig:comapre-citydreamer} presents qualitative results comparing our method to CityDreamer~\cite{city_gen-xie2024citydreamer}, a state-of-the-art (SOTA) 3D city generation pipeline. To ensure a fair comparison, we reproduce CityDreamer by rendering 24 images per scene (its default setting in official implementation) and applying a SOTA reconstruction model~\cite{leroy2024grounding} to extract comparable meshes.
CityDreamer exhibits geometric inconsistencies due to its reliance on image-based rendering, whereas our Sat2City framework achieves significantly improved spatial coherence without compromising appearance quality. Additionally, CityDreamer struggles with prompt adherence, often failing to generate cities that align with the conditional height field and segmentation map. Although CityDreamer can cover a larger area than our method, it remains restricted to limited viewpoints, constraining explicit 3D reconstruction.
Our framework addresses these limitations by directly integrating control conditions into 3D generative priors, enabling precise regulation of building quantities and spatial arrangements that strictly adhere to textual specifications, even without requiring auxiliary inputs such as segmentation maps. Please refer to \textit{Supplementary Material} for more visual results. 


Initial efforts to adapt diffusion models for urban modeling with 3D sparse representations were pioneered by Sat2Scene. While effective at encoding street-level appearance, its representation fails to scale to entire cities. As shown in~\Cref{fig:comapre-sat2scene}, even when provided with ground truth point clouds, Sat2Scene struggles to reconstruct meaningful appearances when trained on our dataset. In contrast, our framework generates high-fidelity urban structures using only noised height-field prompts.
We attribute the scalability limitations of Sat2Scene to its dependence on high point cloud densities: While successful generation of Sat2Scene typically requires approximately 400 points per square meter, our dataset provides only around 14 points per square meter. This fundamental constraint underscores the advantage of our 3D latent sparse representation.

%
%



\begin{table}[t!]
\centering
\setlength{\tabcolsep}{0.4mm}{
\scalebox{0.9}{
\begin{tabular}{lcccc}
\toprule
 & TPQ$\uparrow$ & TSC$\uparrow$ & GPQ$\uparrow$ & GSC$\uparrow$ \\
\midrule
Sat2Scene (2D)~\cite{city_gen-li2024sat2scene} & 6.17 & 5.90 & - & - \\
Sat2Scene (3D)~\cite{city_gen-li2024sat2scene} & 5.57 & 5.47 & 3.83 & 4.05 \\
Sat2Scene* & 3.18 & 3.30 & 3.03 & 3.02 \\
CityDreamer (2D)~\cite{city_gen-xie2024citydreamer} & 6.40 & 6.63 & - & - \\
CityDreamer (3D)* & 4.48 & 4.48 & 3.60 & 3.38 \\
\textbf{Ours} & \textbf{7.35} & \textbf{8.03} & \textbf{6.27} & \textbf{7.02} \\
\bottomrule
\end{tabular}}
}
\vspace{-8pt}
\captionsetup{justification=justified}
\caption{\textbf{Appearance Quality Comparison}. Asterisk (*) denotes methods adapted for fair comparison: Sat2Scene* is retrained on our proposed dataset, and CityDreamer (3D)* leverages MASt3R~\cite{leroy2024grounding} for 3D mesh reconstruction of rendered video sequences. Noted that the Render Quality are directly retrieved from original records.}
\vspace{-12pt}
\label{tab:text_user}
\end{table}

\noindent\textbf{Quantitative Comparison.}
\Cref{tab:geo_eval} demonstrates that our method consistently outperforms well-established scene-level baselines across all geometric metrics. Sat2City achieves a 98.1\% and 7.8\% improvement in COV (CD) and COV (EMD) over previous SOTA, indicating more stable generation quality with a lower probability of mode collapse. Additionally, our method reduces MMD (CD) by 49.4\% and MMD (EMD) by 37.9\% compared to BlockFusion, suggesting superior geometric fidelity and a closer alignment with the reference distribution. It should be noted that the comparison here is between the complete BlockFusion solution and the full Sat2City pipeline, each encompassing both model architecture and associated dataset.

For user perception, \Cref{tab:text_user} shows that Sat2City consistently achieves the highest scores in both textured and geometry-only evaluations. Notably, in metrics evaluating perceived textural quality (TPQ and TSC), our direct 3D generation (7.35 TPQ, 8.03 TSC) surpasses the 2D image rendering-based results of Sat2Scene(2D) (6.17 TPQ, 5.90 TSC) and CityDreamer(2D) (6.40 TPQ, 6.63 TSC). These findings highlight that in large-scale city generation, learning appearance directly from 3D data can also achieve higher user approval than 2D rendering-based methods.

\subsection{Ablations}

\begin{figure}[t!]
  \centering
   \includegraphics[width=0.92\linewidth]{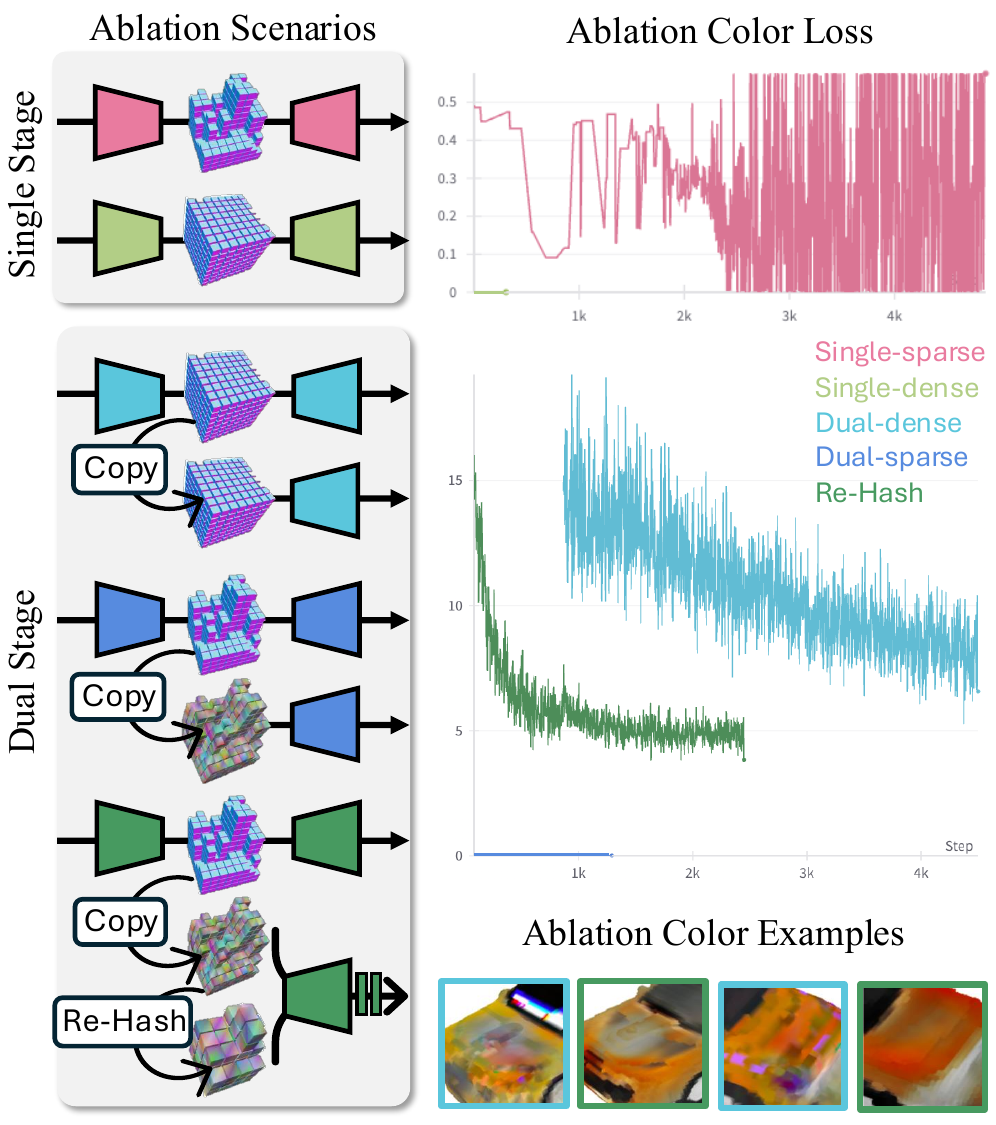}

   \caption{Ablation on \textcolor{customgreen}{Re-Hash} operation. Note that the Single-dense and Dual-sparse variants produce zero values consistently throughout the training process.}
   \label{fig:ablation-main}
   \vspace{-12pt}
\end{figure}

\textbf{Bottleneck Design.} 
The primary objective of our \textcolor{customgreen}{Re-Hash} neck and its dual-stage training is to ensure stable appearance optimization. To validate its effectiveness, we present the color training loss patterns for various bottleneck and training designs in~\Cref{fig:ablation-main} as evidence of our design’s superiority.
Our ablation reveals three key findings: First, encoding both appearance and geometric features into a unified continuous latent representation results in significant gradient direction conflicts, especially for sparse structural configurations. Second, when employing separate latent spaces for color representation through dual-stage training, these gradient conflicts are substantially alleviated for both sparse and dense structures, with the latter beginning to exhibit meaningful gradient descent patterns. 
Notably, the Re-Hash structure demonstrates faster and more stable convergence compared to the dual-dense design, which shows greater loss fluctuation and requires more iterations to stabilize. 
Furthermore, the color rendering results indicates that the dual-dense configuration is prone to generating artifacts in appearance. This deficiency stems from its lack of coarse-grained contextual information, which our Re-Hash structure explicitly incorporates to enable smooth spatial transitions in color representation.


\begin{figure}[t!]
  \centering
   \includegraphics[width=1\linewidth]{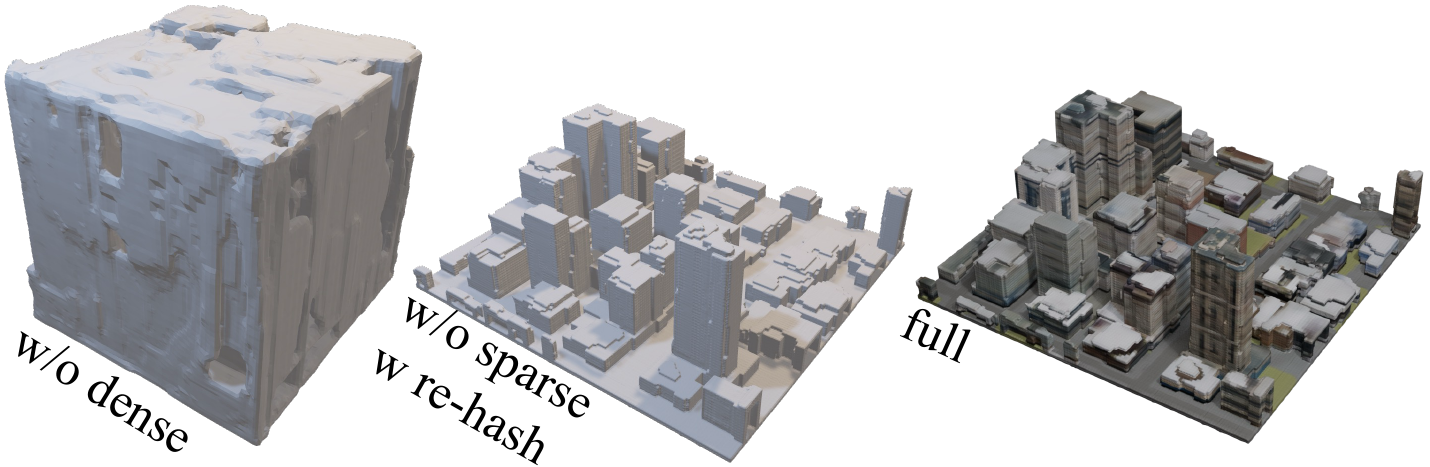}
   \caption{Ablation on cascaded latent diffusion.}
   \label{fig:ablation-diffusion}
   \vspace{-15pt}
\end{figure}

\begin{figure}[t!]
  \centering
   \includegraphics[width=1\linewidth]{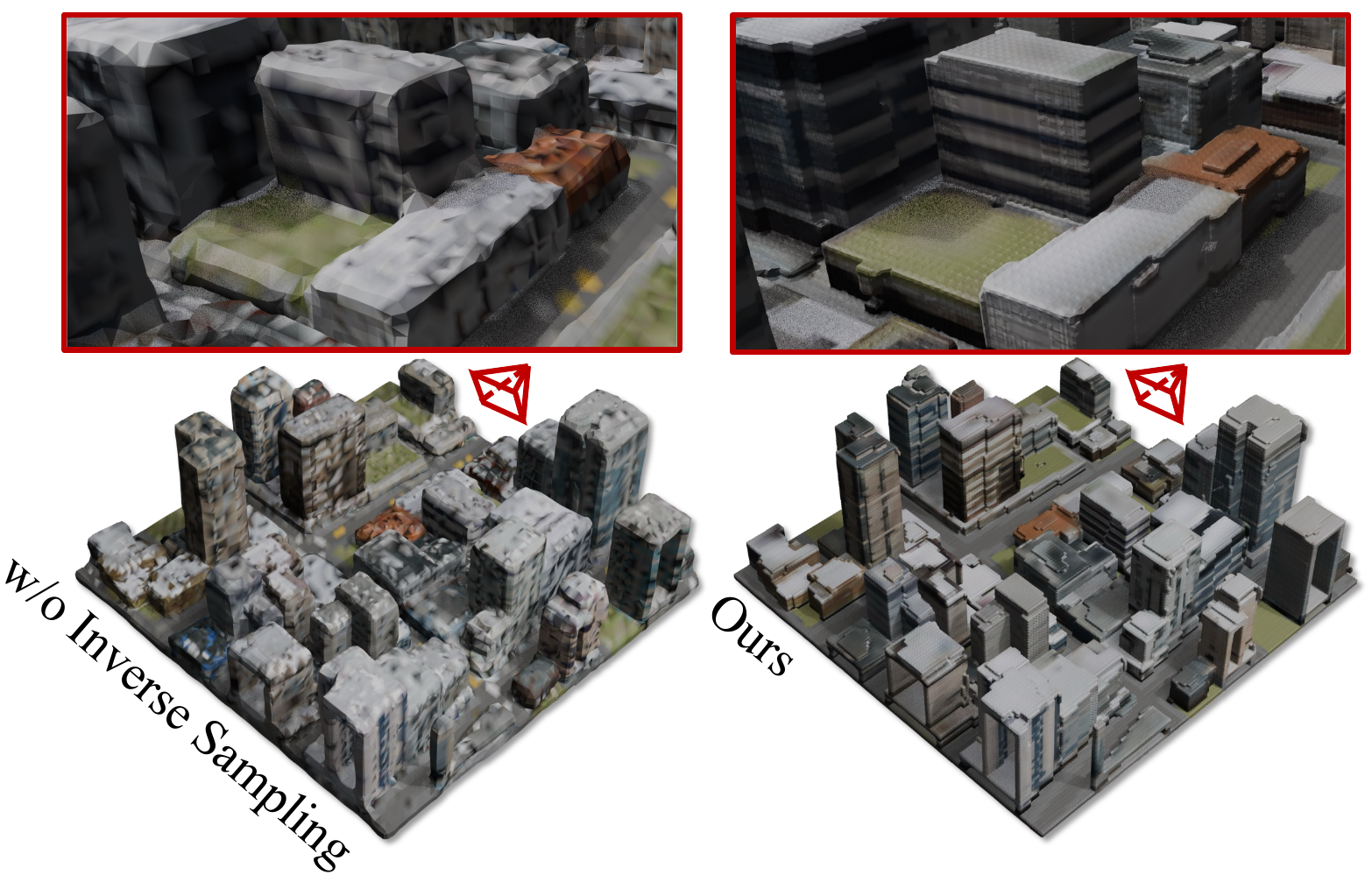}
   \caption{Ablation on inverse sampling.}
   \vspace{-15pt}
   \label{fig:ablate-inverse}
\end{figure}

\noindent\textbf{Inverse-Sampling.} \Cref{fig:ablate-inverse} presents qualitative ablation study results that underscore the importance of our inverse-sampling mechanism. In conventional direct color splatting approaches, where vertex color attributes lack implicit supervision from enclosed voxel points, we observe significant rendering artifacts. Our proposed inverse-sampling enables the color attributes assigned to mesh vertices are continuously constrained by the spatial distribution of points within their associated voxels, which is crucial for maintaining appearance consistency.

\noindent\textbf{Diffusion Model Hierarchy.}  
We ablate each level of our cascaded latent diffusion framework in~\Cref{fig:ablation-diffusion}. Results indicate that relying solely on sparse structures fails to capture unoccupied regions, leading to chaotic generations (w/o dense scenario) under direct sparse latent diffusion. Nevertheless, the sparse latent grid encoder is indispensable for conveying structured pruning decisions essential for appearance decoding (w/o sparse \& w/ re-hash scenario).


\section{Conclusion}
We present Sat2City, a novel framework that integrates sparse voxel grids with latent diffusion models for large-scale 3D city generation. By introducing a cascaded latent diffusion framework with a VAE-based Re-Hash operation and inverse sampling for effectively and efficiently representing appearance. 
Sat2City, trained on our novel dataset of satellite-view simulations and hand-crafted 3D cities, generates detailed 3D urban structures from a single satellite image, surpassing existing models in fidelity. For further discussion of limitations and the potential applicability to real-world satellite–3D Tiles pairs, please refer to the \textit{Supplementary Material}.

{
    \small
    \bibliographystyle{ieeenat_fullname}
    \bibliography{main}
}

\clearpage
\setcounter{page}{1}
\maketitlesupplementary

\section{More Technical details}

\subsection{Implementation Details}
Our method is primarily implemented based on the sparse 3D structure provided by XCube~\cite{ren2024xcube} and the surface fitting network from NKSR~\cite{huang2023neural}. For VAE training, we use 4 Nvidia A800-SXM4-80GB GPUs, while only the diffusion of dense and sparse geometric latent grids is trained across all 4 GPUs; Each layer of the multi-level appearance latent grids is trained on 2 GPUs. The inference process takes approximately \textbf{1 minute} on a single A800 GPU, while the appearanced mesh extraction takes around \textbf{20 seconds} for a resolution of \(512^3\).
The training objectives for the two VAEs—dense geometry and re-hashed appearance—are defined as \(\mathcal{L}_{D}\) and \(\mathcal{L}_{S}\), respectively:
\begin{equation}
\begin{array}{ll}
\mathcal{L}_{D}^{\mathrm{VAE}} & =\mathbb{E}_{\{G,A_N\}}[\mathbb{E}_{X_D\sim q_{\mathcal{E}_d}}[\lambda_0\mathrm{BCE}(G,\tilde{G})+ \\
 & \lambda_1\mathcal{L}_{1}(A_N,\tilde{A}_N)]+ \\&
 
 \lambda_2\mathbb{K}\mathbb{L}(q_{\mathcal{E}_d}(X_D)\parallel p(X_D))],
\\
 \mathcal{L}_{S}^{\mathrm{VAE}} & =\mathbb{E}_{\{G,A_N,P_C\}}[\mathbb{E}_{X_S\sim q_{\mathcal{E}_s}}[\lambda_0\mathrm{BCE}(G,\tilde{G})+ \\
 & \lambda_1\mathcal{L}_{1}(A_N,\tilde{A}_N)+ \lambda_3\mathcal{L}_{1}(P_C,\tilde{P}_C)]+ \\ &\lambda_2\mathbb{K}\mathbb{L}(q_{\mathcal{E}_s}(X_S)\parallel p(X_S))],
\end{array}
\end{equation}where \(\mathrm{BCE}(\cdot)\) represents the binary cross-entropy for grid occupancy, and \(\mathcal{L}_{1}\) denotes the L1 loss. Notably, the KL divergence \(\mathbb{K}\mathbb{L}(\cdot \parallel \cdot)\) is only computed for \(X_D\) and \(X_S\), and not for \(X_{Cn}\). Compared to geometric attributes, appearance attributes often exhibit greater variability and diversity. Introducing KL divergence directly in material learning may overly constrain the model, thus limiting the diversity and flexibility of material features.
The training procedure for 3D latent diffusion follows the same structure for all bottleneck grids using v-parameterization~\cite{salimans2022v-para} and the backbone from~\cite{dhariwal2021diffusion}, with its 3D variant implemented by~\cite{ren2024xcube}.
In practice, we set level \(n=4\), dual training starting epoch as \(E=10\), and the weighting factors are\(\lambda_0 = 20,\lambda_1 = 50,\lambda_2 = 0.03, \lambda_3 = 50\).

\subsection{Details on Cascaded Latent Diffusion}

Our method adopts a cascaded latent diffusion pipeline to generate structured 3D cities with hierarchical geometry and appearance refinement. The pipeline follows a three-stage process: (1) dense geometry latent diffusion, (2) sparse geometry latent refinement, and (3) hierarchical appearance decoding:

\noindent\textbf{Dense latent diffusion conditioned on height field.}  
Inference begins by conditioning on a height map-derived point cloud, which serves as the structural prior for generation. A point encoder first maps the point cloud to voxel indices, extracts features via ResNet blocks, and aggregates them using max-pooling. The unordered point cloud features are then transformed into a structured voxel-based conditioning signal for subsequent encoding. The condition encoder further refines this representation by extracting hierarchical geometric features from sparse voxel grids using multi-scale sparse convolutions. Finally, the encoded condition latent grids are projected to match the resolution of the dense feature grid and concatenated with it for joint iterative denoising. The optimized dense latent diffusion is processed through the dense VAE decoder, generating the first-level sparse grid output (\textit{1st grid}).


\noindent\textbf{Sparse latent diffusion condition on decoded dense latent.}
The sparse latent diffusion serves as a bridge between geometry and appearance representations. It utilizes the \textit{1st grid} to fit a sparse latent volume, facilitating finer-grain surface representation. Crucially, the sparse latent grid decoder records voxel \textit{pruning decisions} at each upsampling step, guiding structured pruning during appearance decoding. This ensures consistency, as the appearance VAE does not explicitly encode geometric distributions.



\noindent\textbf{Re-Hash latent diffusion condition on pruning decision from sparse latent decoder.}
%
%
The \textit{pruning decisions} are subsequently leveraged to guide the upsampling process during appearance decoding, ensuring that the finest-level appearance feature grid maintains a consistent geometric structure with sparse latents at each upsampling step of the appearance decoder. By propagating these decisions, coarser-level appearance features (re-hased from the finest-level) can dynamically adjust their voxel selection strategy based on the most refined masked voxel decisions.  
Specifically, at the bottleneck stage, the finest-level appearance feature grid undergoes a \textit{re-hashing} process, generating multiple coarser levels. During each upsampling step, these coarser levels are iteratively refined by adapting to the pruned voxel structure defined by the finest-level grid. Rather than being naively upsampled, each coarser level is re-fitted to the structural surface, ensuring alignment with the progressively pruned geometry at that specific stage of upsampling. This hierarchical conditioning ensures spatial coherence and tightly constrains the appearance latent around the defined surfaces.

Once the geometry structure is established, the multi-level appearance latents are decoded by performing trilinear interpolation over the inferred geometry vertices. The interpolated latent features across all levels are concatenated and processed through an MLP-based output head to generate per-vertex appearance attributes, ensuring spatial consistency across the generated 3D scene.

\paragraph{Diffusion model with parameter $\Psi$.} A widely adopted forward diffusion step (adding noise) is given by:

\begin{equation}
    \mathbf{X}_t | \mathbf{X}_{t-1} \sim \mathcal{N}(\sqrt{1 - \beta_t} \mathbf{X}_{t-1}, \beta_t \mathbf{I}),
\end{equation}where the noise variance $\beta_t$ is small, ensuring gradual corruption of data over $T$ diffusion steps. The reverse process attempts to denoise $\mathbf{X}_t$ progressively, ultimately restoring the original data distribution. It is parameterized as:

\begin{equation}
    \mathbf{X}_{t-1} | \mathbf{X}_t \sim \mathcal{N}(\boldsymbol{\mu}_{\Psi}(\mathbf{X}_t, t), \frac{1 - \bar{\alpha}_{t-1}}{1 - \bar{\alpha}_t} \beta_t \mathbf{I}),
\end{equation}where $\alpha_t = 1 - \beta_t$ and $\bar{\alpha}_t = \prod_{s=0}^{t} \alpha_s$. The mean $\boldsymbol{\mu}_{\Psi}$ is parameterized by a learnable neural network. In practice, we re-express $\boldsymbol{\mu}_{\Psi}$ using:

\begin{equation}
    \boldsymbol{\mu}_{\Psi} = \sqrt{\alpha_t} \mathbf{X}_t - \beta_t \sqrt{\frac{\bar{\alpha}_{t-1}}{1 - \bar{\alpha}_t}} \mathbf{v}_{\Psi},
\end{equation}where the network is trained to predict $\mathbf{v}$ instead of the noise, following the v-parameterization strategy, which has been found to improve optimization stability. Following XCube, the $v_{\Psi}(\cdot)$ is tailored for our sparse representations based on the one that originally proposed for dense image space~\cite{dhariwal2021diffusion}.

\begin{figure}[t!]
  \centering
   \includegraphics[width=1\linewidth]{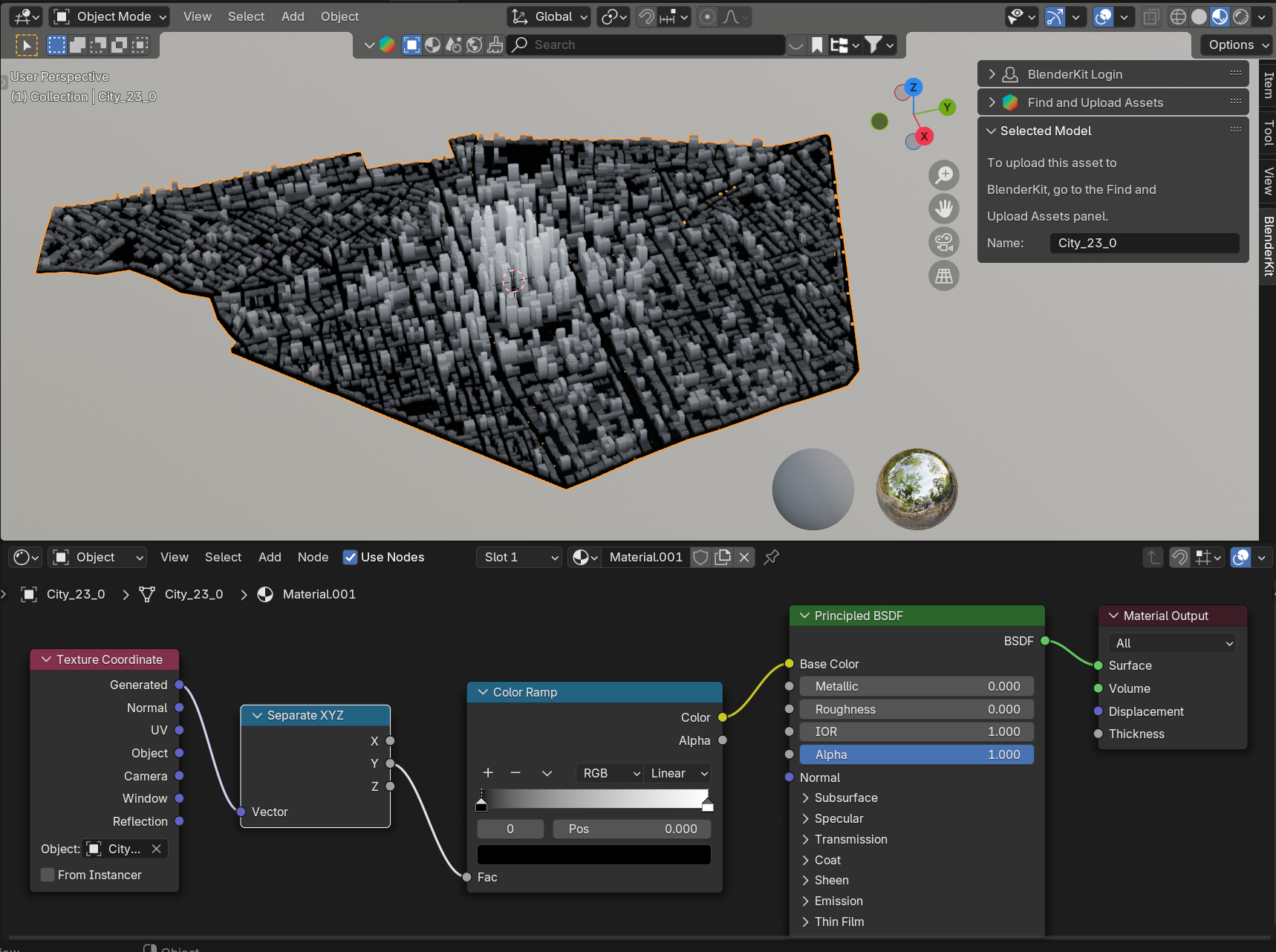}
   \caption{Height Map Rendering in Blender.}
   \label{fig:hmap-blender}
   \vspace{-12pt}
\end{figure}

\subsection{Details on Dataset Height Map Construction}

We employ a procedural shading technique using a height-dependent grayscale gradient (\Cref{fig:hmap-blender}). The shading process is implemented within the Blender shading node, leveraging a combination of procedural texture coordinates, channel separation, and color mapping.

\noindent\textbf{Coordinate-Based Height Extraction}
We begin by extracting the generated texture coordinates of the city mesh using the Texture Coordinate node. This provides a spatially varying reference frame that adapts dynamically to the geometry. The coordinate vector is subsequently decomposed into its individual components  (x, y, z)  through a Separate XYZ node, isolating the vertical height  z  for use in the shading computation.

\noindent\textbf{Height-Driven Color Mapping}
The extracted height information is passed through a Color Ramp node, which maps height values to a grayscale color gradient. This mapping is crucial for simulating height-based shading variations, where lower buildings are rendered darker, and higher structures receive lighter intensities, mimicking ambient occlusion and environmental exposure effects.

\noindent\textbf{Physically-Based Material Composition}
The resulting color gradient is then fed into the Base Color input of a Principled BSDF shader. This shader governs the material’s light interaction properties, maintaining a physically consistent representation of urban structures. For enhanced realism, the roughness is set to 1.0 (diffuse reflection), and metallic properties are disabled to simulate non-reflective building surfaces.


\section{Discussion}


\noindent\textbf{Limitations.} The evaluation of our approach on real-world datasets remains pending, as obtaining high-resolution colorized point clouds precisely aligned with remote sensing elevation maps necessitates considerable resource allocation and may encounter limitations in data availability and dissemination. 

\begin{figure*}[t!]
  \centering
  \includegraphics[width=1\linewidth]{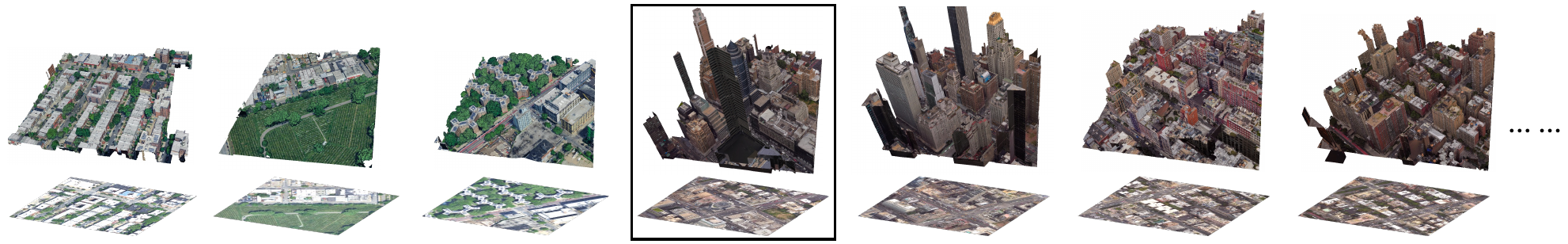}
   \caption{Real-world dataset includes diverse morphologies.}
   \label{fig:long-row}
\end{figure*}

\noindent\textbf{Spoiler alert.} We developed an automated pipeline leveraging Google Earth Engine and the Google Maps Platform to retrieve (i) high-resolution satellite imagery and (ii) co-registered photorealistic 3D Tiles for identical geographic bounding boxes. \Cref{fig:long-row} shows several such samples, and we have already begun constructing a large-scale dataset.

We evaluated the data processing workflow—comprising API-based data acquisition, point cloud conversion, and height map inference—on several commercially available desktop PCs. The complete pipeline requires approximately 10 minutes per scene, indicating that large-scale data collection is practically achievable within a standard academic laboratory environment.

\begin{figure}[t!]
  \centering
  \includegraphics[width=1\linewidth]{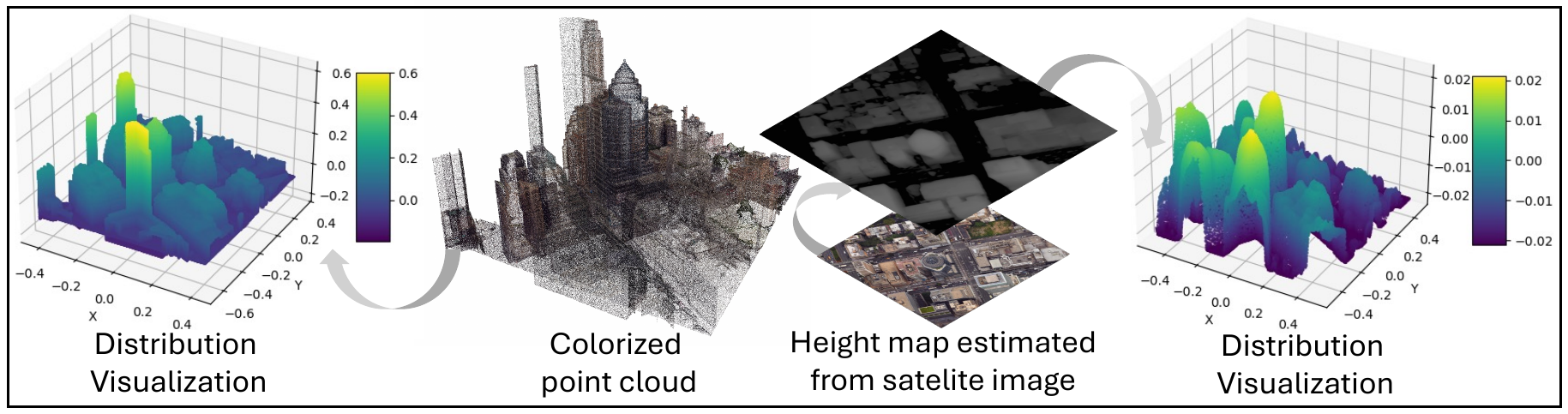}
   \caption{Raw data example for one scene in~\Cref{fig:long-row}.}
   \label{fig:one}
\end{figure}

As an initial step toward evaluating Sat2City’s generalizability, we fine-tuned Depth Anything~\cite{yang2024depth} on a dataset comprising real-world height maps and corresponding satellite imagery~\cite{xiong2023gamus}. We inferred the scenes shown in \Cref{fig:long-row}.
\underline{Quantitatively}, we computed the Chamfer Distance between point clouds derived from height maps and their colorized ground truth, yielding 0.2909 for synthetic data (Sat2City dataset) and 0.0977 for real-world data (Google Earth), both averaged over seven randomly selected scenes.
\underline{Qualitatively}, \Cref{fig:one} (real) illustrates that the height map distribution in the real-world dataset exhibits greater overall consistency with the colorized point clouds compared to its synthetic counterpart shown in \Cref{fig:dataset-hmap} (synthetic).
These results suggest that real-world data is unlikely to introduce generalization problems and may even improve the quality of generation.

\noindent\textbf{Future works.} Although contributing an additional real-world dataset is beyond the scope of this submission, as concurrent works (SynCity~\cite{engstler2025syncity} and NuiScene~\cite{lee2025nuiscene}) likewise validate only on synthetic data, we plan to explore it in future work fully. Nevertheless, our findings underscore both the necessity and the promise of building real-world datasets to push this research direction forward. We also hope our work will inspire a series of follow-up studies in the remote sensing community to explore the model’s limitations and possible improvements on such real-world dataset—examining, for example, its performance in various urban environments (\eg, diverse terrains, or mixed vegetation and building layouts), at different remote sensing resolutions.

\section{More About Evaluations}
\subsection{User Study Settings}

\Cref{tab:user-study-design} presents our user study design, where participants were instructed to rate images based on the provided questions. The evaluation follows a 10-point scale, where 1 indicates “Very Poor” and 10 represents “Excellent.” For each sample, participants answered two questions—either texture-based or geometry-based—ensuring a comprehensive assessment of both perceptual quality and structural fidelity.

\begin{table}[ht!]
\small
    \centering
    \renewcommand{\arraystretch}{1.2} 
    \begin{tabularx}{\linewidth}{lX}
        \hline
        \textbf{Metric} & \textbf{User Study Question} \\
        \hline
        TPQ & How would you rate how the scene looks overall? Think about the details, the textures, and how realistic it seems to you. \\
        TSC & How well do you think the shapes and structure of the scene are represented? Does the geometry look accurate and complete? \\
        GPQ & How would you rate how the scene looks overall? Think about the details, the textures, and how realistic it seems to you. \\
        GSC & How well do you think the shapes and structure of the scene are represented? Does the geometry look accurate and complete? \\
        \hline
    \end{tabularx}
    \caption{User study metrics and evaluation questions.}
    \label{tab:user-study-design}
\end{table}

\subsection{Texture Evaluation.} Our evaluation of the generated textures is limited to qualitative visual analysis, primarily because this work does not involve any neural rendering components that produce 2D images. Moreover, standard 2D generation metrics such as FID or KID are not applicable for assessing cross-modal outputs—namely, the input colorized point cloud and the resulting textured mesh.

\subsection{More Dataset \& Generation Snapshots}

\begin{figure}[t!]
  \centering
   \includegraphics[width=1\linewidth]{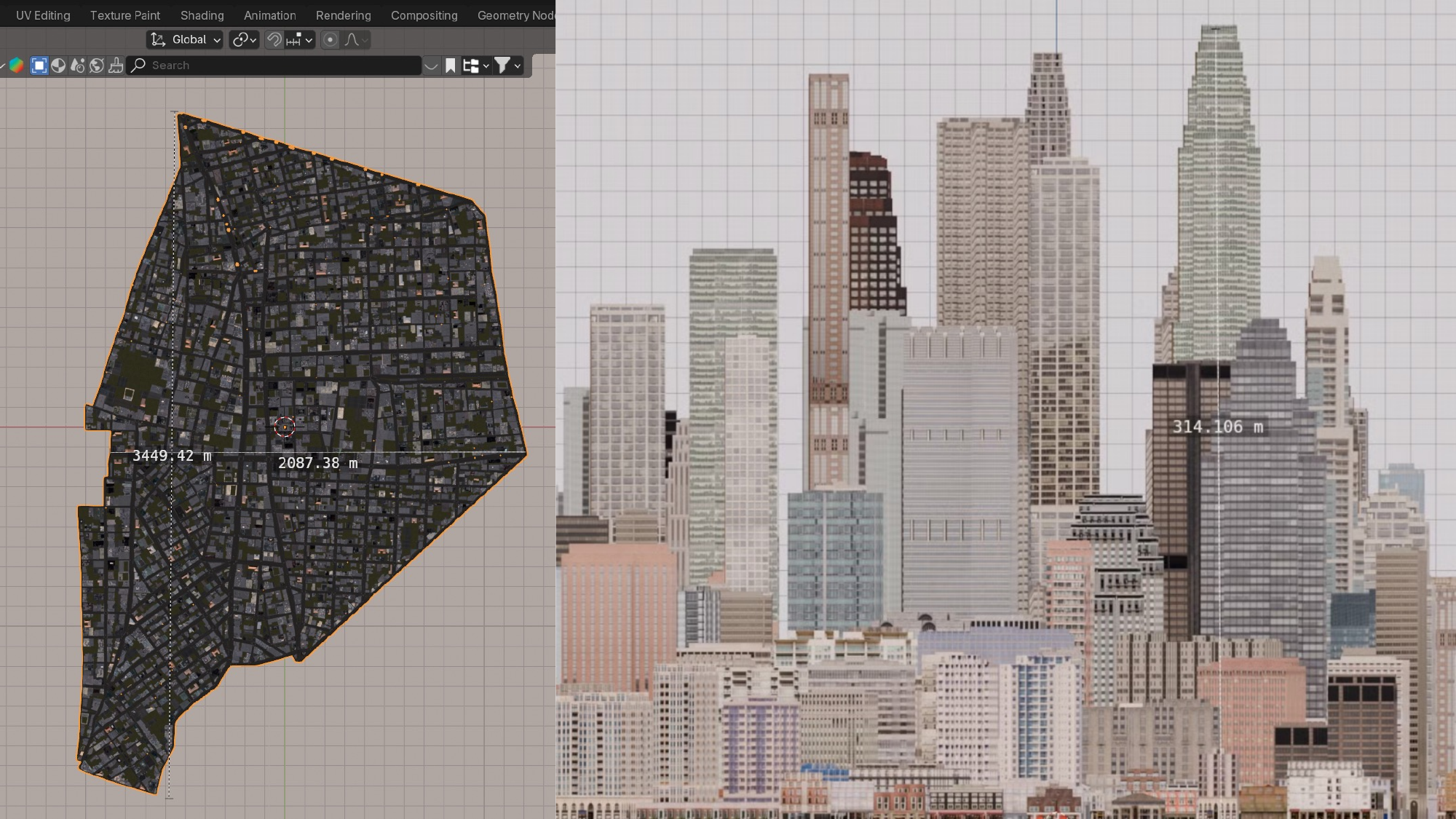}
   \caption{Orthographic camera view for synthetic satellite imagery and building distribution.}
   \label{fig:hmap-blender}
\end{figure}

\begin{figure*}[t!]
  \centering
   \includegraphics[width=1\linewidth]{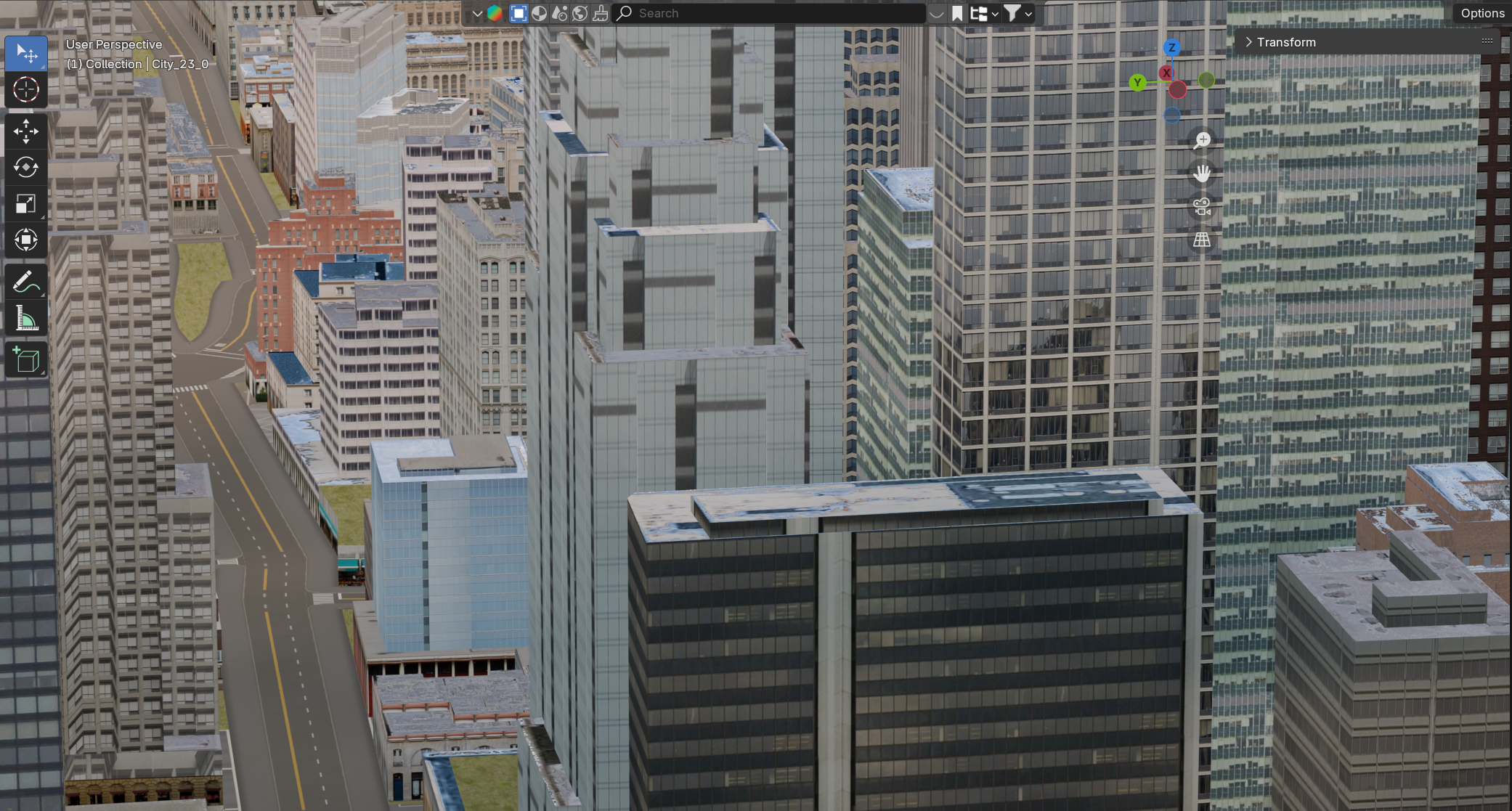}
\caption{Exploring the original artist-designed mesh model.}
   \label{fig:hmap-blender}
\end{figure*}

\begin{figure*}[t!]
  \centering
   \includegraphics[width=1\linewidth]{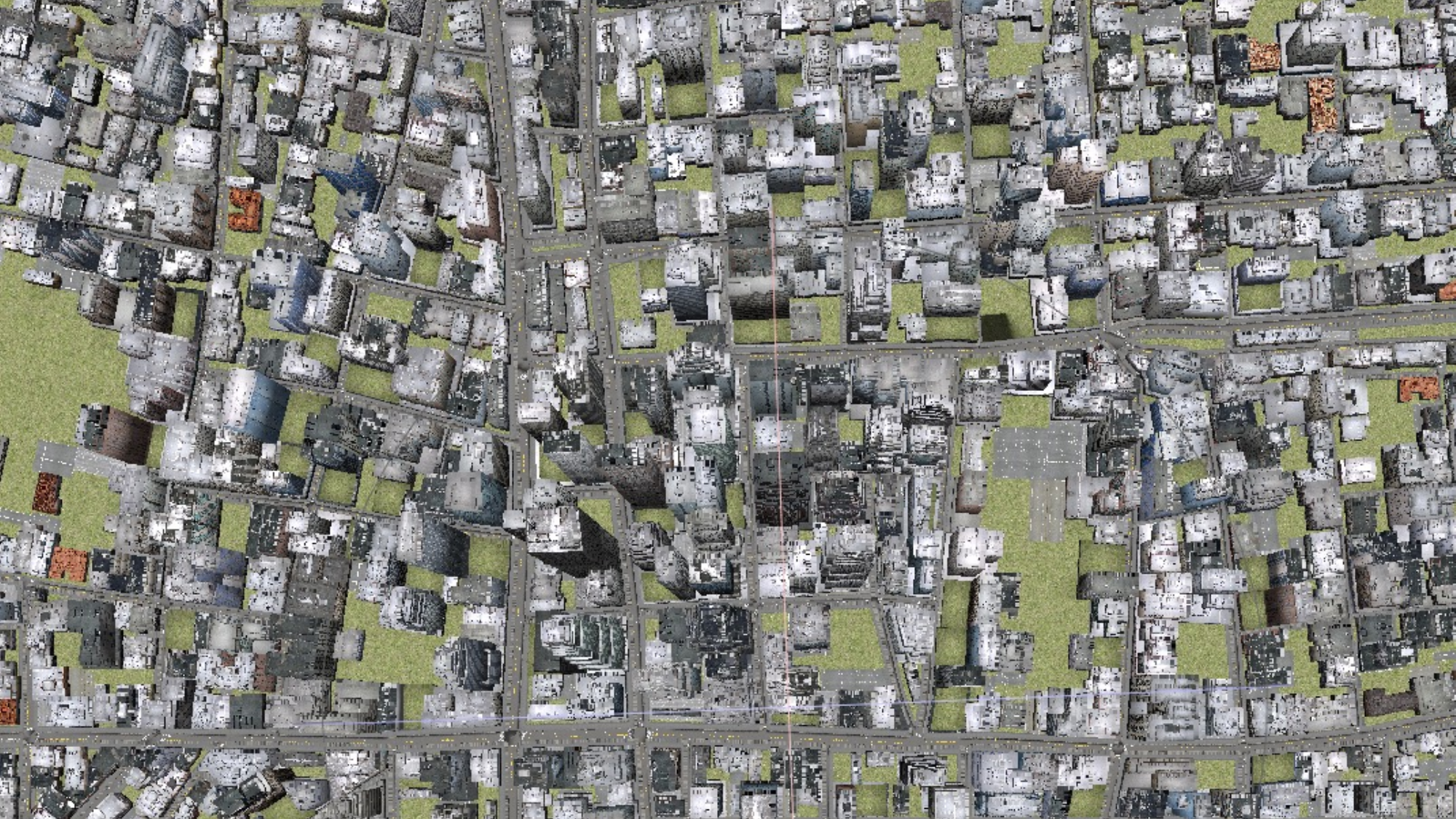}
\caption{More views on entire sampled point cloud.}
   \label{fig:pc-zoomout}
\end{figure*}

\begin{figure*}[t!]
  \centering
   \includegraphics[width=1\linewidth]{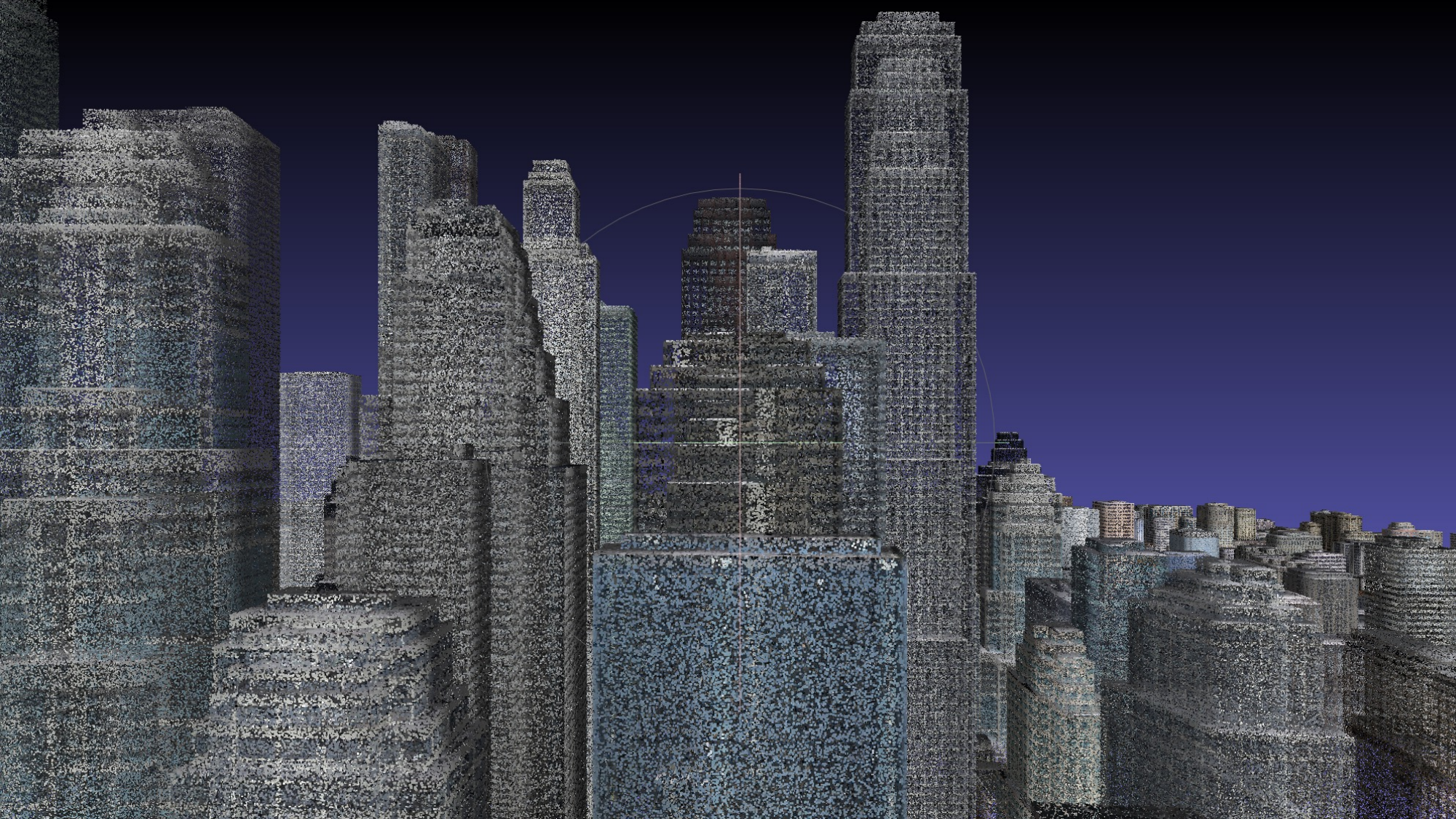}
\caption{More views on zoomed-in sampled point cloud.}
   \label{fig:pc-zoomin}
\end{figure*}

\begin{figure*}[t!]
  \centering
   \includegraphics[width=1\linewidth]{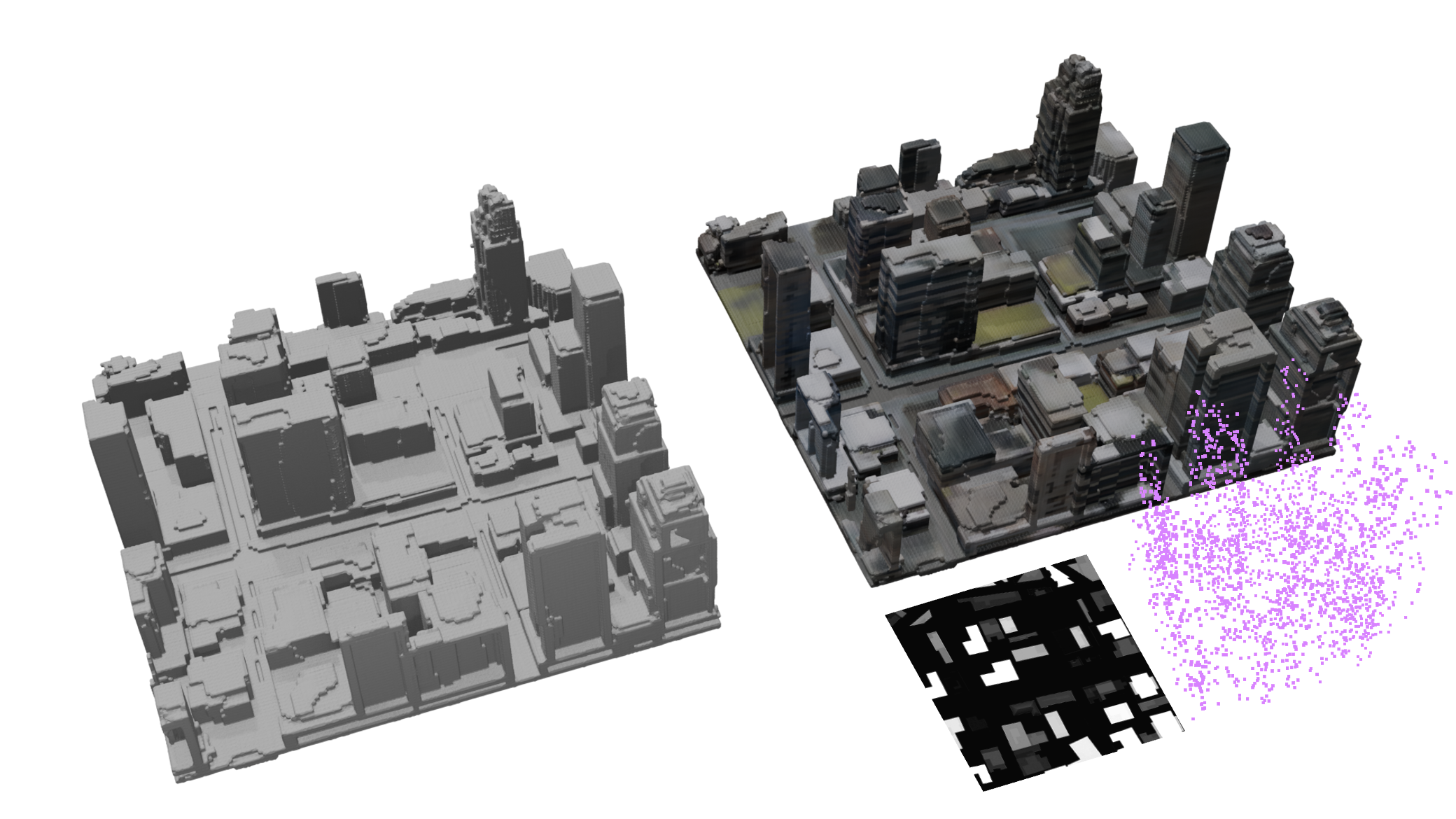}
   \label{fig:pc-zoomin}
\end{figure*}

\begin{figure*}[t!]
  \centering
   \includegraphics[width=1\linewidth]{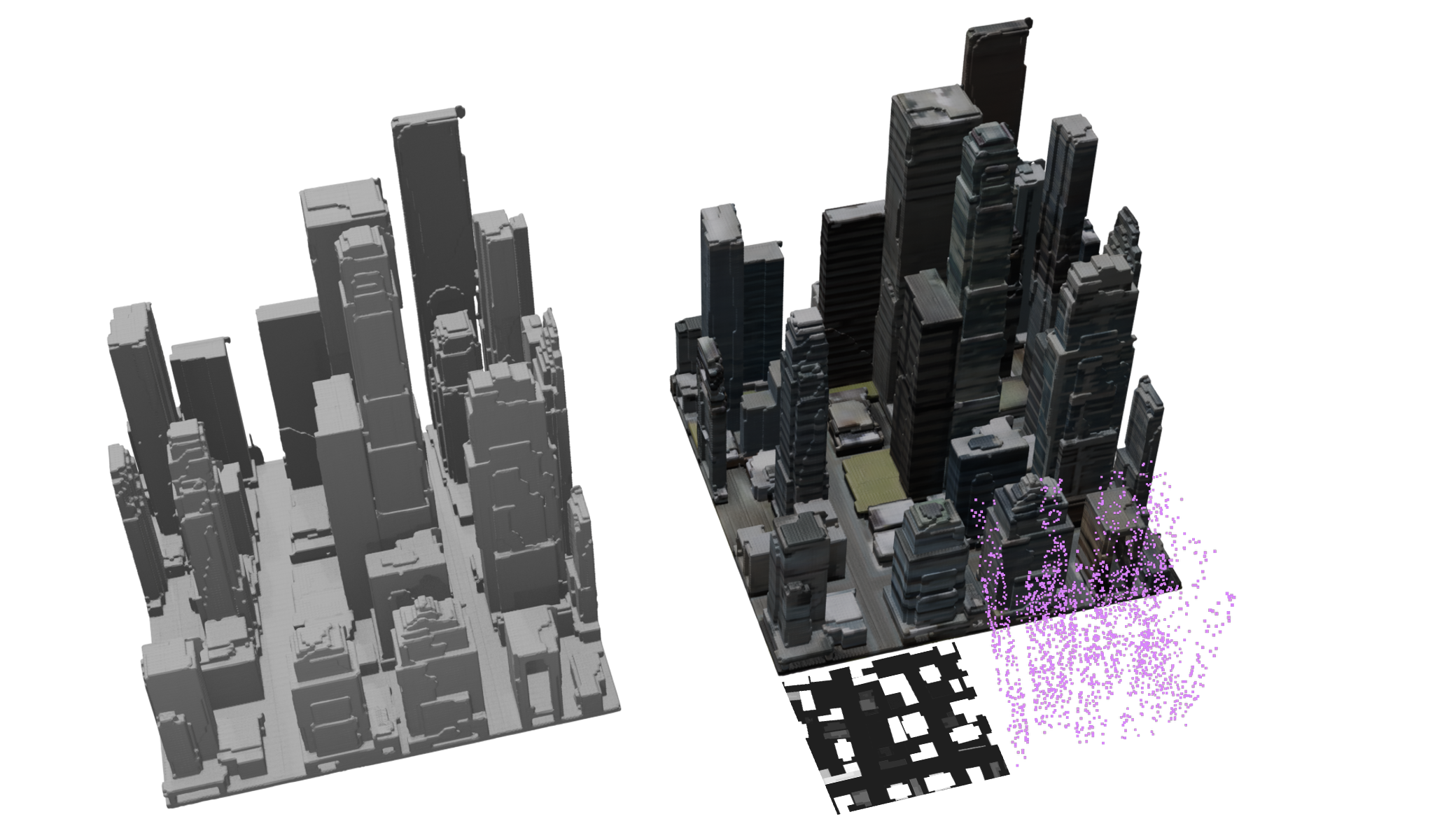}
   \label{fig:pc-zoomin}
\end{figure*}

\begin{figure*}[t!]
  \centering
   \includegraphics[width=1\linewidth]{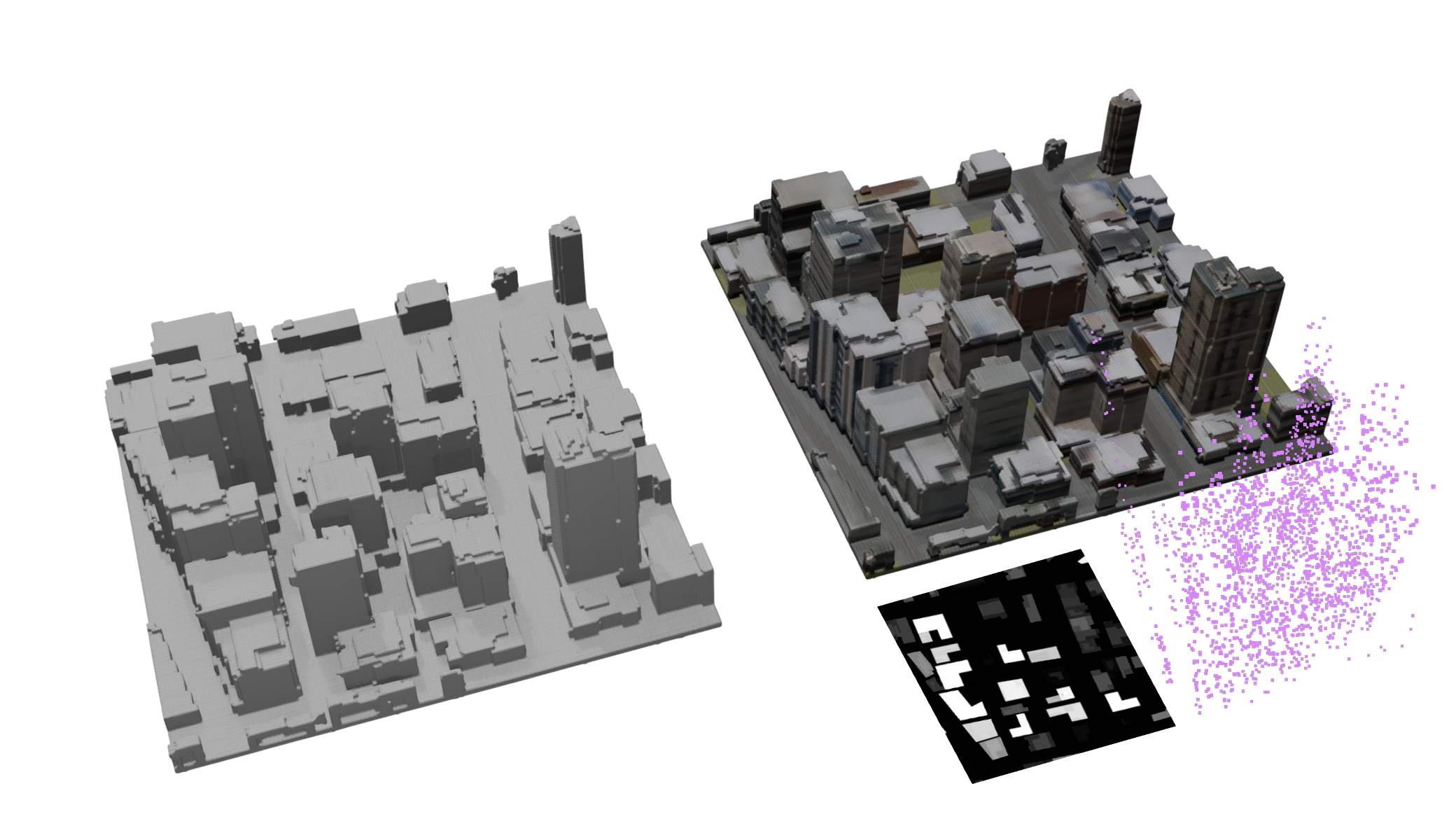}
\caption{More results.}
   \label{fig:pc-zoomin}
\end{figure*}


\end{document}